\documentclass[10pt,twocolumn,letterpaper]{article}

\usepackage{iccv}
\usepackage{times}
\usepackage{epsfig}
\usepackage{graphicx}
\usepackage{amsmath}
\usepackage{amssymb}

\usepackage{algorithm}
\usepackage{algorithmicx}
\usepackage{algpseudocode}
\usepackage{color}
\usepackage{enumitem}
\usepackage{arydshln}
\usepackage{multirow}
\usepackage{bm}
\usepackage{authblk}

\usepackage[breaklinks=true,bookmarks=false]{hyperref}

\iccvfinalcopy 

\def\iccvPaperID{4305} 

\ificcvfinal\fi

\begin{document}

\title{\scalebox{0.9}{Efficient and Accurate Arbitrary-Shaped Text Detection with Pixel Aggregation Network}}

\author[1]{Wenhai Wang$^*$}
\author[2,4]{Enze Xie\thanks{Authors contributed equally.}}
\author[1]{Xiaoge Song}
\author[3]{Yuhang Zang}
\author[2]{Wenjia Wang}
\author[1]{Tong Lu\thanks{Corresponding author.}}
\author[4]{Gang Yu}
\author[5]{Chunhua Shen}
\affil[1]{National Key Lab for Novel Software Technology, Nanjing University}
\affil[2]{Tongji University}
\affil[3]{University of Electronic Science and Technology of China}
\affil[4]{Megvii (Face++) Technology Inc.}
\affil[5]{The University of Adelaide}
\affil[ ]{\tt\small\{wangwenhai362, Johnny\_ez, sxg514\}@163.com, yuhangzang@foxmail.com, wwj940312@126.com\\
\tt\small lutong@nju.edu.cn, yugang@megvii.com, chunhua.shen@adelaide.edu.au}

\maketitle
\ificcvfinal\thispagestyle{empty}\fi

\begin{abstract}
Scene text detection, an important step of scene text reading systems, has witnessed rapid development with convolutional neural networks. Nonetheless, two main challenges still exist and hamper its deployment to real-world applications. The first problem is the trade-off between speed and accuracy. The second one is to model the arbitrary-shaped text instance. Recently, some methods have been proposed to tackle arbitrary-shaped text detection, but they rarely take the speed of the entire pipeline into consideration, which may fall short in practical applications.
In this paper, we propose an efficient and accurate arbitrary-shaped text detector, termed Pixel Aggregation Network (PAN), which is equipped with a low computational-cost segmentation head and a learnable post-processing. More specifically, the segmentation head is made up of Feature Pyramid Enhancement Module (FPEM) and Feature Fusion Module (FFM). FPEM is a cascadable U-shaped module, which can introduce
%
multi-level information to guide the better segmentation. FFM can gather the features given by the FPEMs of different depths into a final feature for segmentation. The learnable post-processing is implemented by Pixel Aggregation (PA), which can precisely aggregate text pixels by predicted similarity vectors. Experiments on several standard benchmarks validate the superiority of the proposed PAN. It is worth noting that our method can achieve a competitive F-measure of 79.9\% at 84.2 FPS on CTW1500.

%
\end{abstract}

\begin{figure}[t]
		\centering
		\setlength{\fboxrule}{0pt}
		\fbox{\includegraphics[width=0.45\textwidth]{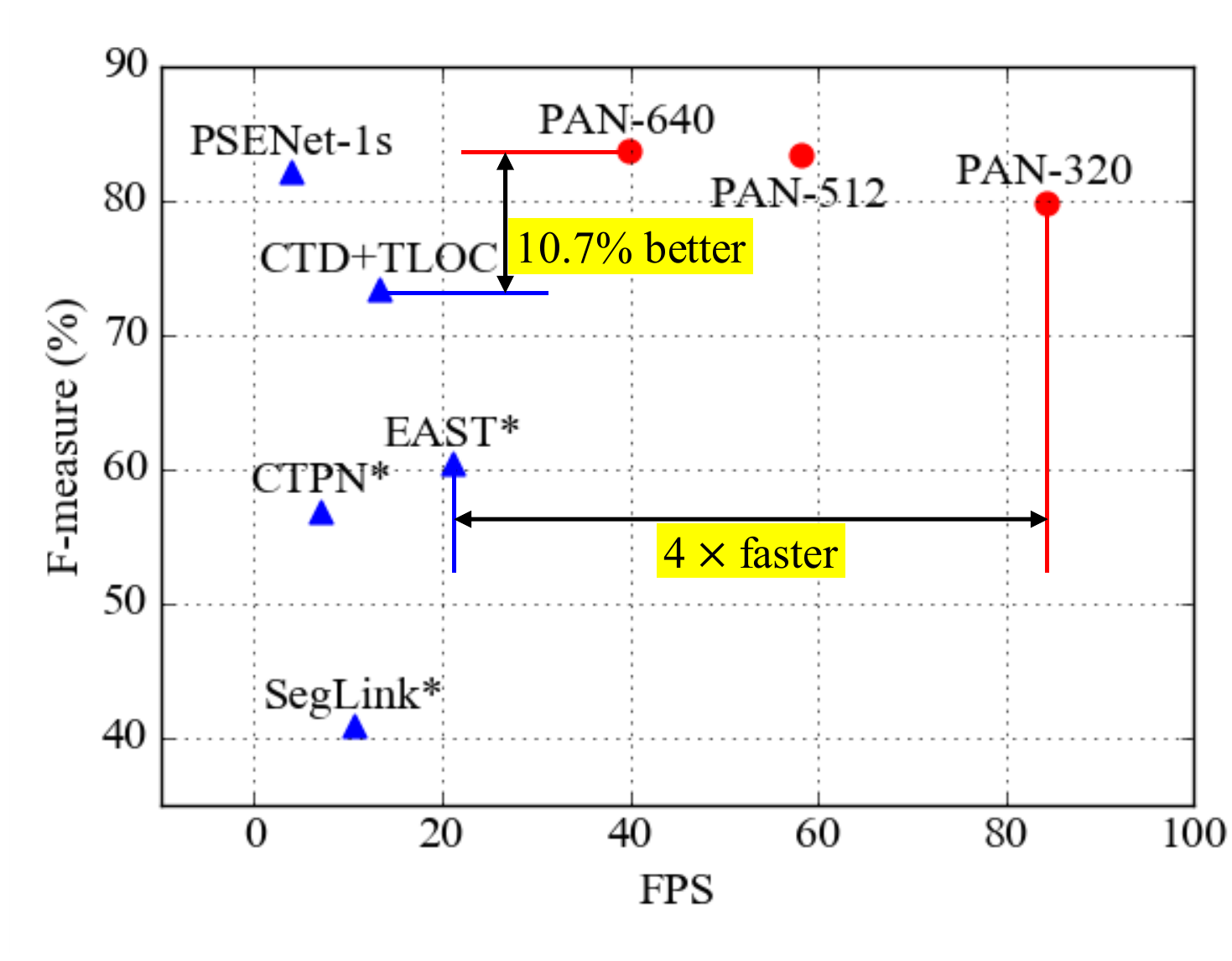}}
		\caption{The performance and speed on curved text dataset CTW1500. 
		PAN-640 is 10.7\% better than CTD+TLOC, and PAN-320 is 4 times faster than EAST.
		* indicates the results from \cite{Liu2017Detecting}.}
		\label{fig:f_fps}
		\vspace{-4pt}
\end{figure}
\section{Introduction}
Scene text detection is a fundamental and critical task in computer vision,
as it is a key step in many text-related applications, such as text recognition, text retrieval, license plate recognition and text visual question answering.
In virtue of recent development of object detection~\cite{ren2015faster,liu2016ssd,he2017mask,fan2019shifting,fan2018salient,fan2019rethinking,zhao2019contrast} and segmentation \cite{chen2017rethinking,zhao2017pyramid,long2015fully,yu2018bisenet,fan2017structure,fan2018enhanced} based on CNN~\cite{he2016identity,huang2017densely,wang2018mixed,li2019selective}, scene text
detection has witnessed great progress~\cite{tian2016detecting,shi2017detecting,zhou2017east,Liu2017Detecting,textsnake,spcnet,psenet}.
Arbitrary-shaped text detection, one of the most challenging tasks in text detection,
%
is receiving
more and more research  attention
Some new methods~\cite{Liu2017Detecting,textsnake,spcnet,psenet} have been put forward to detect curve text instance. However, many of these methods
suffer from low inference speed, because of their heavy models or complicated post-processing steps, which limits their deployment in the real-world environment. On the other hand, previous text detectors~\cite{zhou2017east,mcn} with high efficiency are mostly designed for quadrangular text instances, which have flaws when detecting curved text. Therefore, ``how to design an efficient and accurate arbitrary-shaped text detector'' remains 
largely unsolved.
%


%

\begin{figure}[t]
		\centering
		\setlength{\fboxrule}{0pt}
		\fbox{\includegraphics[width=0.47\textwidth]{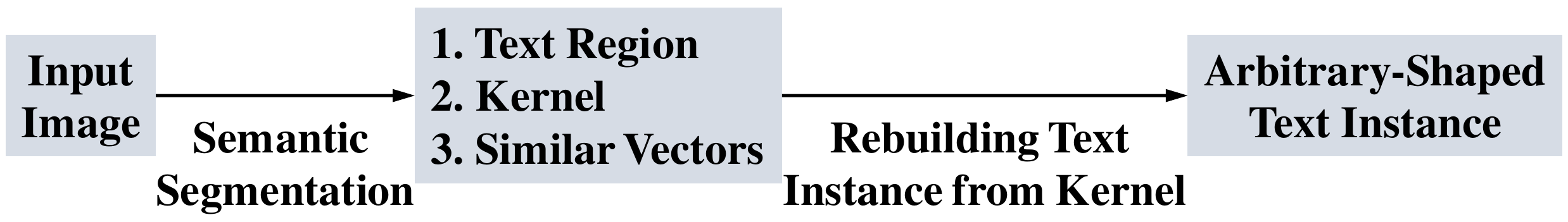}}
		\caption{The
		overall
		pipeline of PAN.
		}
		\label{fig:pipeline}
		\vspace{-4pt}
\end{figure}

\begin{figure*}[t]
	\centering
	\setlength{\fboxrule}{0pt}
	\fbox{\includegraphics[width=0.9\textwidth]{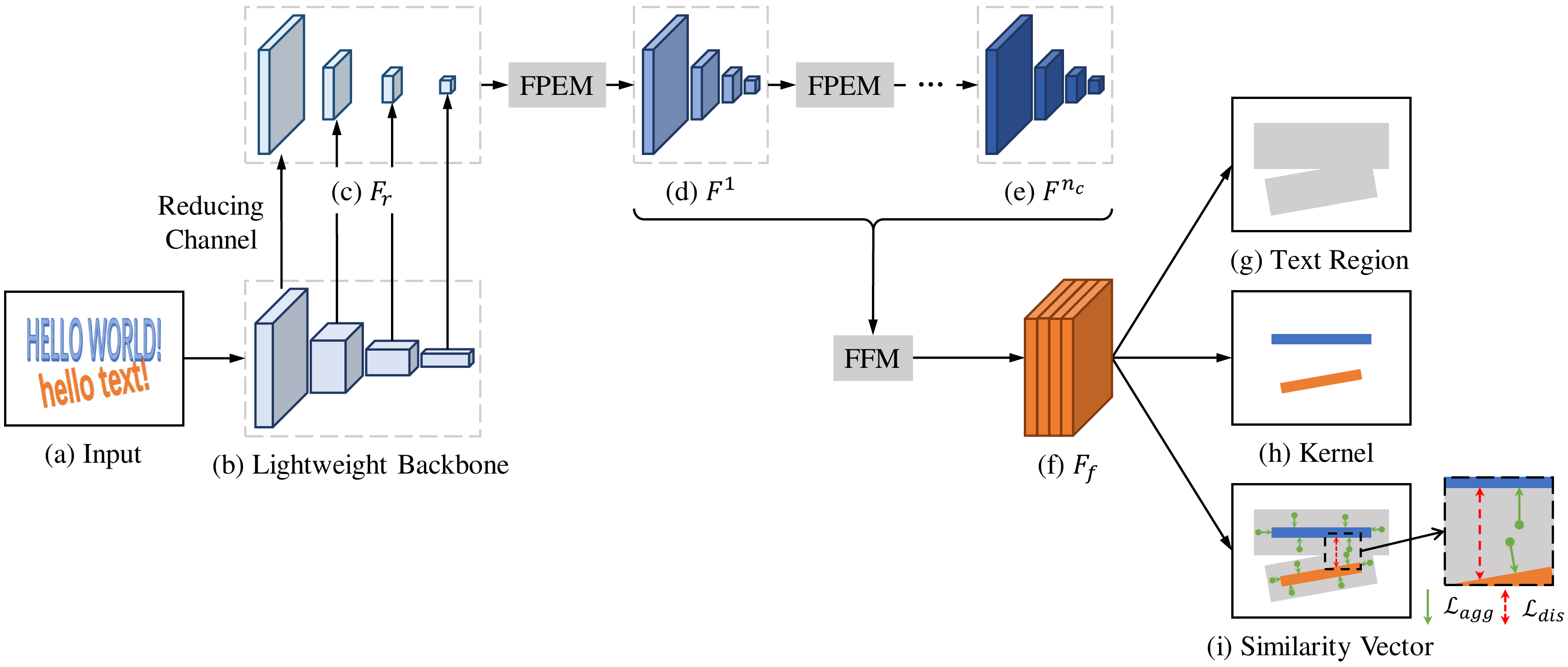}}
	\caption{The overall architecture of PAN. The features from lightweight backbone network are enhanced by a low computational-cost segmentation head which is composed of Feature Pyramid Enhancement Module (FPEM) and Feature Fusion Module (FFM). The network predicts text regions, kernels and similarity vectors to describe the text instances.
	}
	\label{fig:arch}
	\vspace{-4pt}
\end{figure*}

To solve these problems,
here we propose an arbitrary-shaped text detector, namely Pixel Aggregation Network (PAN), which can achieve a good balance between speed and performance.
PAN makes  arbitrary-shaped text detection following the simple pipeline as shown in Fig.~\ref{fig:pipeline},
which only contains two steps: \romannumeral1) Predicting the text regions, kernels and similarity vectors by segmentation network. \romannumeral2) Rebuilding complete text instances from the predicted kernels.
For high efficiency, we need to reduce the time cost of these two steps. First and foremost, a lightweight backbone is required for segmentation. In this paper, we use ResNet18~\cite{he2016deep} as the default backbone of PAN.
However, the lightweight backbone is relatively weak in feature extraction, and thus its features typically have small receptive fields and
weak
representation capabilities. To remedy this defect, we
propose
a low computation-cost segmentation head, which is composed of two modules: Feature Pyramid Enhancement Module (FPEM) and Feature Fusion Module (FFM).
FPEM is a U-shaped module built by separable convolutions (see Fig.~\ref{fig:fpem}), and therefore FPEM is able to enhance the features of different scales by fusing the low-level and high-level information with
minimal computation overhead.
Moreover, FPEM is cascadable, which allows us to compensate for the depth of lightweight backbone by appending FPEMs after it (see Fig.~\ref{fig:arch}~(d)(e)).
To gather low-level and high-level semantic information, before final segmentation, we introduce FFM to fuse the features generated by the FPEMs of different depths.
In addition, to reconstruct complete text instances accurately, we propose a learnable post-processing method, namely Pixel Aggregation (PA), which can guide the text pixels to correct kernels through the predicted similarity vectors.

To show the effectiveness of our proposed PAN, we conduct extensive experiments on four challenging benchmark datasets including CTW1500~\cite{Liu2017Detecting}, Total-Text~\cite{totaltext}, ICDAR 2015~\cite{karatzas2015icdar} and MSRA-TD500~\cite{msra}. Among these datasets, CTW1500 and Total-Text are new datasets designed for curve text detection.
As shown in Fig.~\ref{fig:f_fps}, on CTW1500, the F-measure of PAN-640 is 83.7\% which is 10.7\% better than CTD+TLOC~\cite{Liu2017Detecting}, and the FPS of PAN-320 is 84.2 which is 4 times faster than EAST~\cite{zhou2017east}.
Meanwhile, PAN also has promising performance on multi-oriented and long text datasets.

In summary, our contributions are three-fold.
Firstly, we propose a lightweight segmentation neck consisting of Feature Pyramid Enhancement Module (FPEM) and Feature Fusion Module (FFM) which are two high-efficiency modules that can improve the feature representation of the network.
Secondly, we propose Pixel Aggregation (PA), in which the text similarity vector can be learned by the network and be used to selectively aggregate pixels nearby the text kernels.
Finally, the proposed method achieves state-of-the-art performance on two curved text benchmarks while still keeping the inference speed of 58 FPS.
To our 
knowledge, 
ours is the first algorithm which can detect curved text precisely in real-time.

\section{Related Work}
In recent years, text detectors based on deep learning have achieved remarkable results.
Most of these methods can be roughly divided into two categories:
anchor-based methods and anchor-free methods.
Among these methods, some use a heavy framework or complicated pipeline for high accuracy,
while others adopt a simple structure to maintain a good balance between speed and accuracy.

\textbf{Anchor-based text detectors}
are usually inspired by object detectors such as Faster R-CNN~\cite{ren2015faster} and SSD~\cite{ren2015faster}.
TextBoxes~\cite{liao2017textboxes} directly modifies the anchor scales and shape of convolution kernels of SSD to handle text with extreme aspect ratios.
TextBoxes++~\cite{textboxes++} further regresses quadrangles instead of horizontal bounding boxes for multi-oriented text detection.
RRD~\cite{rrd} applies rotation-invariant and sensitive features for text classification and regression from two separate branches for better long text detection.
SSTD~\cite{he2017single} generates text attention map to enhance the text region of the feature map and suppress background information, which is beneficial for tiny texts.
Based on Faster R-CNN, RRPN~\cite{rrpn} develops rotated region proposals to detect titled text.
Mask Text Spotter~\cite{masttextspotter} and SPCNet~\cite{spcnet} regard text detection as an instance segmentation problem and
use
Mask R-CNN~\cite{maskrcnn} for arbitrary text detection.
The above-mentioned methods
achieve remarkable results on several benchmarks. Nonetheless, most of them rely on complex anchor setting, which makes these approaches heavy-footed and prevent them from
applying to
real-world problems.

\textbf{Anchor-free text detectors}
formulate text detection as a text segmentation problem, which are often
built upon
fully convolutional networks (FCN)~\cite{FCN}.
Zhang et al.~\cite{zhang2016multi} first estimate
text blocks with FCNs and detect
character candidates from those text blocks with MSER.
Yao et al.~\cite{yao2016scene} use FCN to predict three parts of a text instance, text/non-text, character classes, and character linking orientations, then apply a group process for text detection.
To separate adjacent text instances, PixelLink~\cite{PixelLink} performs text/non-text and links prediction in pixel level, then applies  post-processing to obtain text boxes and excludes noises.
EAST~\cite{zhou2017east} and DeepReg~\cite{deepreg} adopt FCNs to predict shrinkable text score maps and perform per-pixel regression, followed by a post-processing NMS.
TextSnake~\cite{textsnake} models text instances with ordered disks and text center lines, which is able to represent text in arbitrary shapes.
PSENet~\cite{psenet} uses FCN to predict text instances with multiple scales, then adopts progressive scale expansion algorithm to reconstruct the whole text instance. Briefly speaking, the main differences among anchor-free methods are the way of text label generation and post-processing. Nevertheless, among these methods, only TextSnake and PSENet are designed for detecting curved text instances which also widely appear in natural scenes. However, they suffer from a heavy framework or a complicated pipeline, which usually slows down their inference speed.

\textbf{Real time text detection} requires a fast way to generate high-quality text prediction. EAST~\cite{zhou2017east} directly 
employs
FCNs to predict the score map and corresponding coordinates and,  followed by a simple NMS. The whole pipeline of EAST is concise so that it can maintain a relatively high speed.
MCN~\cite{mcn} 
formulates
text detection problem as a graph-based clustering problem and generates bounding boxes without using NMS, which can be fully parallelized on GPUs.
However, these methods are designed for quadrangular text detection and fail to locate the curve text instances.


\section{Proposed Method}

\subsection{Overall Architecture}
PAN follows a segmentation-based pipeline (see Fig.~\ref{fig:pipeline}) to detect arbitrary-shaped text instances.
For high efficiency, the backbone of the segmentation network
must be lightweight.
However, the features
offered
by a lightweight backbone often have small receptive fields and weak
representation capabilities.
For this reason, we propose  segmentation head that is computationally efficient to refine the features. The segmentation head contains two key modules, namely Feature Pyramid Enhancement Module~(FPEM) and Feature Fusion Module~(FFM).
As  shown in Fig.~\ref{fig:arch}~(d)(e) and Fig.~\ref{fig:fpem}, FPEM is cascadable and has low computational cost, which
can be attached behind the backbone to make features of different scales deeper and more expressive.
After that, we employ the Feature Fusion Module~(FFM) to fuse the features produced by the FPEMs of different depths into a final feature for segmentation.
PAN predicts text regions~(see Fig.~\ref{fig:arch}~(g)) to describe the complete shapes of text instances, and predicts kernels (see Fig.~\ref{fig:arch}~(h)) to distinguish different text instances.
The network also predicts similarity vector~(see Fig.~\ref{fig:arch}~(i)) for each text pixel, so that the distance between the similarity vectors of pixel and kernel from the same text instance is small.

Fig.~\ref{fig:arch}
shows
the  overall architecture of PAN. We employ a lightweight model~(ResNet-18~\cite{he2016deep}) as the backbone network of PAN.
There are 4 feature maps~(see Fig.~\ref*{fig:arch}~(b)) generated by conv2, conv3, conv4, and conv5 stages of backbone, and note that they have strides of 4, 8, 16, 32 pixels with respect to the input image.
We use 1$\times$1 convolution to reduce the channel number of each feature map to 128, and get a thin feature pyramid $F_r$. The feature pyramid is enhanced by $n_c$ cascaded FPEMs. Each FPEM produces an enhanced feature pyramid, and thus there are $n_c$ enhanced feature pyramids $F^1$, $F^2$,..., $F^{n_c}$. FFM fuses the $n_c$ enhanced feature pyramids into a feature map $F_f$, whose stride is 4 pixels and the channel number is 512. $F_f$ is used to predict text regions, kernels and similarity vectors. Finally, we apply a simple and efficient post-processing algorithm to obtain the final text instances.

\subsection{Feature Pyramid Enhancement Module}
\begin{figure}[t]
		\centering
		\setlength{\fboxrule}{0pt}
		\fbox{\includegraphics[width=0.45\textwidth]{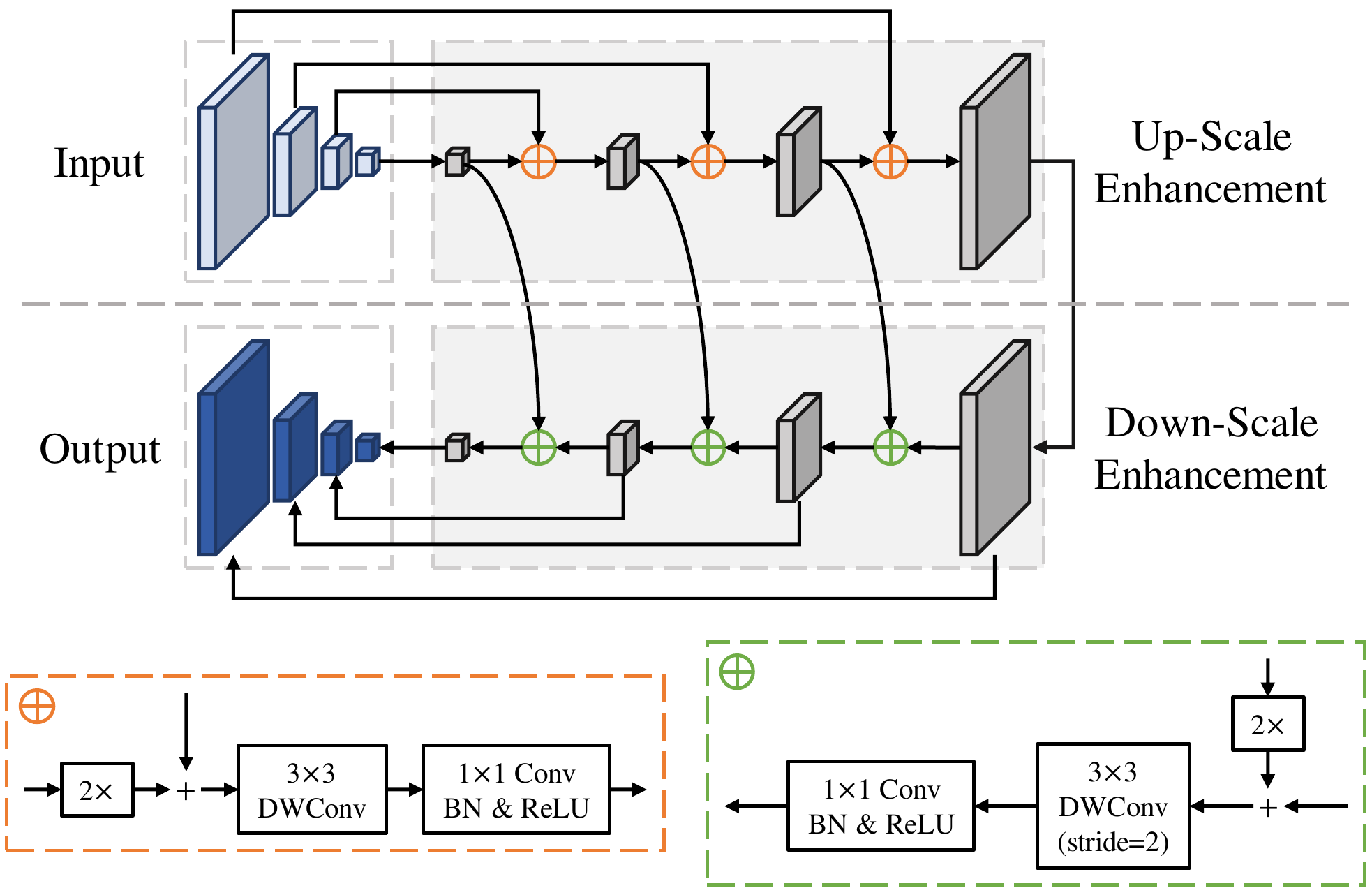}}
		\caption{The details of FPEM. ``$+$'', ``2$\times$'', ``DWConv'', ``Conv'' and ``BN'' represent element-wise addition, 2$\times$ linear upsampling, depthwise convolution~\cite{howard2017mobilenets}, regular convolution~\cite{lecun1998gradient} and Batch Normalization~\cite{ioffe2015batch} respectively.}
		\label{fig:fpem}
		\vspace{-4pt}
\end{figure}

FPEM is a U-shaped module as illustrated in Fig.~\ref{fig:fpem}. It consists of two phases,
namely,
up-scale enhancement and down-scale enhancement.
The up-scale enhancement acts on the input feature pyramid.
In this phase, the enhancement is iteratively performed on the feature maps with strides of 32, 16, 8, 4 pixels.
In the down-scale phase, the input is the feature pyramid generated by up-scale enhancement, and the enhancement is conducted from 4-stride to 32-stride.

Meanwhile, the output feature pyramid of down-scale enhancement is the final output of FPEM.
We employ separable convolution~\cite{howard2017mobilenets} (3$\times$3 depthwise convolution~\cite{howard2017mobilenets} followed by 1$\times$1 projection) instead of the regular convolution to build the join part $\oplus$ of FPEM (see the dashed frames in Fig.~\ref{fig:fpem}).
Therefore, FPEM is capable of enlarging the receptive field (3$\times$3 depthwise convolution) and deepening the network (1$\times$1 convolution) with a small computation overhead.

Similar to FPN~\cite{lin2017feature}, FPEM is able to enhance the features of different scales by fusing the low-level and high-level information.
In addition, different from FPN, there are two other advantages of FPEM.
Firstly, FPEM is a cascadable module. With the increment of cascade number $n_c$, the feature maps of different scales are fused more adequately and the receptive fields of features become larger.
Secondly,
FPEM is computationally cheap.
FPEM is built by separable convolution, which needs
minimal
computation. The FLOPS of FPEM is about $1/5$ of FPN.

\subsection{Feature Fusion Module}
\label{sec:ffm}
\begin{figure}[t]
		\centering
		\setlength{\fboxrule}{0pt}
		\fbox{\includegraphics[width=0.45\textwidth]{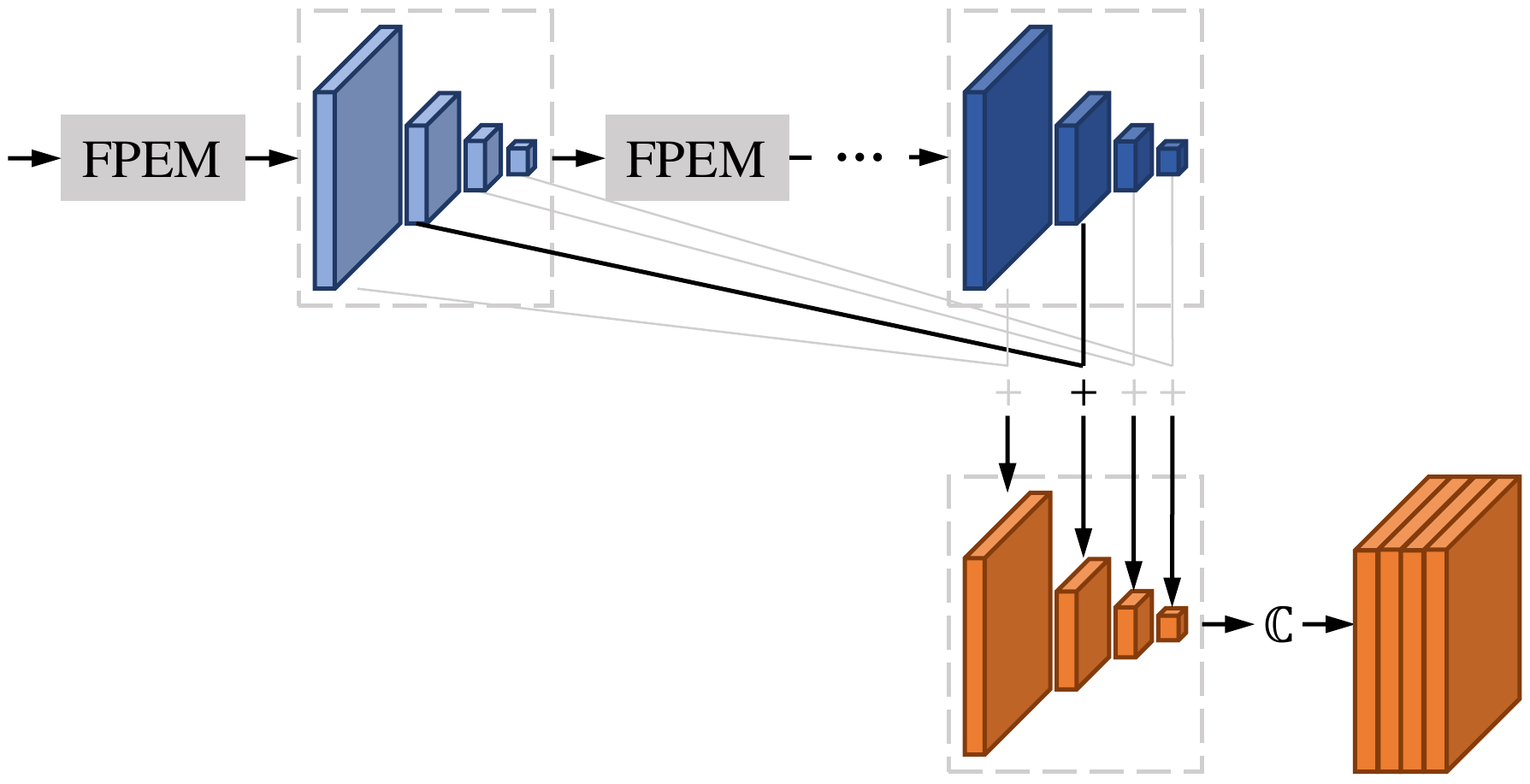}}
		\caption{The detail of FFM. ``$+$'' is element-wise addition. ``$\mathcal{C}$'' is the operation of upsampling and concatenating.}
		\label{fig:fpfm}
		\vspace{-4pt}
\end{figure}

Feature Fusion Module is applied to fuse the feature pyramids $F^1$, $F^2$,..., $F^{n_c}$ of different depths. Because both low-level and high-level semantic information are important for semantic segmentation.
A direct and effective method to combine these feature pyramids is to upsample and concatenate them. However, the fused feature map given by this method has a large channel number (4$\times$128$\times n_c$), which slows down the final prediction. Thus, we propose another fusion method as shown in Fig.~\ref{fig:fpfm}.
We firstly combine the corresponding-scale feature maps by element-wise addition. Then, the feature maps after addition are upsampled and concatenated into a final feature map
which only has 4$\times$128 channels.

\subsection{Pixel Aggregation}
The text regions keep the complete shape of text instances, but the text regions of the text instances lying closely are often overlapping (see Fig.~\ref{fig:arch}~(g)).
Contrarily, the text instances can be well distinguished using the kernels (see Fig.\ref{fig:arch}~(h)). However, the kernels are not the complete text instance. To rebuild the complete text instances, we need to merge the pixels in text regions to kernels.
We propose a learnable algorithm, namely Pixel Aggregation, to guide the text pixels towards correct kernels.

In Pixel Aggregation, we borrow the idea of clustering to reconstruct the complete text instances from the kernels.
Let us consider the text instances as clusters. The kernels of text instances are cluster centers. The text pixels are the samples to be clustered.
Naturally, to aggregate the text pixels to the corresponding kernels, the distance between the text pixel and kernel of the same text instance should be small. In the training phase, we use aggregation loss $\mathcal{L}_{agg}$ as Equ.~\ref{eqn:loss_agg} to implement this rule.
\begin{equation}
	\mathcal{L}_{agg} = \frac{1}{N} \sum_{i=1}^{N} \frac{1}{\left| T_i \right|} \sum_{p \in T_i} \mathop{ln}(\mathcal{D}(p, K_i) + 1),
	\label{eqn:loss_agg}
\end{equation}
\begin{equation}
	\mathcal{D}(p, K_i) = max(\left\| \mathcal{F}(p) - \mathcal{G}(K_i) \right\| - \delta_{agg}, 0)^2,
	\label{eqn:dis_pk}
\end{equation}
where the $N$ is the number of text instances. The $T_i$ is the $i$th text instance. ${D}(p, K_i)$ defines the distance between text pixel $p$ and the kernel $K_i$ of text instance $T_i$. $\delta_{agg}$ is a constant, which is set to 0.5 experimentally and used to filter easy samples. $\mathcal{F}(p)$ is the similarity vector of the pixel $p$. $\mathcal{G}(\cdot)$ is the similarity vector of the kernel $K_i$, which can be calculated by $\sum_{q \in K_i}\mathcal{F}(q) / \left| K_i \right|$.

In addition, the cluster centers need to keep discrimination. Therefore, the kernels of different text instances should maintain enough distance. We use discrimination loss $\mathcal{L}_{dis}$ as Equ.~\ref{eqn:loss_dis} to describe this rule during the training.
\begin{equation}
	\mathcal{L}_{dis} = \frac{1}{N(N-1)} \sum_{i=1}^{N} \mathop{\sum_{j=1}^{N}}\limits_{j\neq i} \mathop{ln}(\mathcal{D}(K_i, K_j) + 1),
	\label{eqn:loss_dis}
\end{equation}
\begin{equation}
	\mathcal{D}(K_i, K_j) = \mathop{max}(\delta_{dis} - \left\| \mathcal{G}(K_i) - \mathcal{G}(K_j) \right\|, 0)^2.
	\label{eqn:dis_kk}
\end{equation}
$\mathcal{L}_{dis}$ try to keep the distance among the kernels not less than $\delta_{dis}$ which is set to 3 in all our experiments.

In the testing phase, we use the predicted similarity vector to guide the pixels in the text area to the corresponding kernel. The detailed post-processing steps are as follows: \romannumeral1) Finding the connected components in the kernels' segmentation result, and each connected component is a single kernel. \romannumeral2) For each kernel $K_i$, conditionally merging its neighbor text pixel (4-way) $p$ in predicted text regions while the Euclidean distance of their similarity vectors is less than $d$.
\romannumeral3) Repeating step \romannumeral2) until there is no eligible neighbor text pixel.

\subsection{Loss Function}

Our loss function can be formulated as:
\begin{equation}
	\mathcal{L} =  \mathcal{L}_{tex} + \alpha \mathcal{L}_{ker} + \beta (\mathcal{L}_{agg} + \mathcal{L}_{dis}),
	\label{eqn:loss-tot}
\end{equation}
where $\mathcal{L}_{tex}$ is the loss of the text regions and $\mathcal{L}_{ker}$ is the loss of the kernels.
The $\alpha$ and $\beta$ are used to balance the importance among $\mathcal{L}_{tex}$, $\mathcal{L}_{ker}$, $\mathcal{L}_{agg}$ and $\mathcal{L}_{dis}$, and we set them to 0.5 and 0.25 respectively in all experiments.

Considering the extreme imbalance of text and non-text pixels, we follow \cite{psenet} and adopt dice loss~\cite{milletari2016v} to supervise the segmentation result $P_{tex}$ of the text regions and $P_{ker}$ of the kernels.
Thus $\mathcal{L}_{tex}$ and $\mathcal{L}_{ker}$ can be written as follows:
\begin{equation}
	\mathcal{L}_{tex} = 1 - \frac{2 \sum_{i} P_{tex}(i)G_{tex}(i)}{\sum_i P_{tex}(i)^2 + \sum_i G_{tex}(i)^2},
	\label{eqn:loss_tex}
\end{equation}
\begin{equation}
	\mathcal{L}_{ker} = 1 - \frac{2 \sum_{i} P_{ker}(i)G_{ker}(i)}{\sum_i P_{ker}(i)^2 + \sum_i G_{ker}(i)^2},
	\label{eqn:loss_ker}
\end{equation}
where $P_{tex}(i)$ and $G_{tex}(i)$ refer to the value of the $i$th pixel in the segmentation result and the ground truth of the text regions respectively. The ground truth of the text regions is a binary image, in which text pixel is 1 and non-text pixel is 0. Similarly, $P_{ker}(i)$ and $G_{ker}(i)$ means the $i$th pixel value in the prediction and the ground truth of the kernels. The ground truth of the kernels is generated by shrinking original ground truth polygon, and we follows the method in \cite{psenet} to shrink the original polygon by ratio $r$.
Note that, we adopt Online Hard Example Mining (OHEM)~\cite{shrivastava2016training} to ignore simple non-text pixels when calculating ${L}_{tex}$, and we only take the text pixels in ground truth into consideration while calculating $\mathcal{L}_{ker}$, $\mathcal{L}_{agg}$ and $\mathcal{L}_{dis}$.

\section{Experiment}

\subsection{Datasets}
\textbf{SynthText}~\cite{synthtext} is a large scale synthetically generated dataset containing 800K synthetic images. Following \cite{shi2017detecting,lyu2018multi,textsnake}, we pre-train our model on this dataset.

\textbf{CTW1500}~\cite{Liu2017Detecting} is a recent challenging dataset for curve text detection. It has 1000 training images and 500 testing images. The dataset focus on curve text instances which are labeled by 14-polygon.

\textbf{Total-Text}~\cite{totaltext} is also a newly-released dataset for curve text detection. This dataset includes horizontal, multi-oriented and curve text instances and consists of 1255 training images and 300 testing images.

\textbf{ICDAR 2015} (IC15)~\cite{karatzas2015icdar} is a commonly used dataset for text detection. It contains a total of 1500 images, 1000 of which are used for training and the remaining are for testing. The text instances are annotated by 4 vertices of the quadrangle.

\textbf{MSRA-TD500} includes 300 training images and 200 test images with text line level annotations. It is a dataset with multi-lingual, arbitrary-oriented and long text lines. Because the training set is rather small, we follow the previous works \cite{zhou2017east,lyu2018multi,textsnake} to include the 400 images from HUST-TR400~\cite{yao2014unified} as training data.

\subsection{Implementation Details}
We use the ResNet~\cite{he2016identity} or VGG16~\cite{simonyan2014very} pre-trained on ImageNet~\cite{deng2009imagenet} as our backbone. The dimension of the similarity vector is set to 4.
All the networks are optimized by using stochastic gradient descent (SGD).
The pre-trained model is trained on SynthText for 50K iterations with a fixed learning rate of $1\times10^{-3}$.
Two training strategies are adopted in other experiments: \romannumeral1) Training from scratch. \romannumeral2) Fine-tuning on SynthText pre-trained model.
When training from scratch, we train PAN with batch size 16 on 4 GPUs for 36K iterations, and the initial learning rate is set to $1\times10^{-3}$. Similar to \cite{zhao2017pyramid}, we use the ``poly'' learning rate strategy in which the initial rate is multiplied by $(1 - \frac{iter}{max\_iter})^{power}$, and the $power$ is set to 0.9 in all experiments.
When fine-tuning on SynthText pre-trained model, the number of iterations is 36K, and the initial learning rate is $1\times10^{-3}$.
We use a weight decay of $5 \times 10^{-4}$ and a Nesterov momentum~\cite{sutskever2013importance} of 0.99. We adopt the weight initialization introduced by~\cite{he2015delving}.

In the training phase, we ignore the blurred text regions labeled as “DO NOT CARE” in all datasets. The negative-positive ratio of OHEM is set to 3. We apply random scale, random horizontal flip, random rotation and random crop on training images. On ICDAR 2015 and MSRA-TD500, we fit a minimal area rectangle for each predicted text instance. The shrink ratio $r$ of the kernels is set to 0.5 on ICDAR 2015 and 0.7 on other datasets. In the testing phase, the distance threshold $d$ is set to 6.


\subsection{Ablation Study}
To make the conclusion of ablation studies more generalized, all experiments of ablation studies are conducted on ICDAR 2015 (a quadrangle text dataset) and CTW1500 (a curve text dataset). Note that, in these experiments, all models are trained without any external dataset. The short sides of test images in ICDAR 2015 and CTW1500 are set to 736 and 640 respectively.

\textbf{The influence of the number of cascaded FPEMs.} We study the effect of the number of cascaded FPEMs by varying $n_c$ from $0$ to $4$. Note that, when $n_c = 0$, we upsample and concatenate the feature maps in $F_r$ to get $F_f$.
From Table~\ref{tab:abs_nc}, we can find that the F-measures on the test sets keep rising with the growth of $n_c$ and begins to level off when $n_c \ge 2$.
However, a large $n_c$ will slow down the model despite the low computational cost of FPEM. For each additional FPEM, the FPS will decrease by about 2-5 FPS.
To keep a good balance of performance and speed, we set $n_c$ to 2 by default in the following experiments.

\begin{table}[t]
	\scriptsize
	\centering
	\renewcommand\arraystretch{1}
	\newcommand{\tabincell}[3]{\begin{tabular}{@{}#1@{}}#2\end{tabular}}
	\scalebox{1}{
		\begin{tabular}{|c|c|c|c|c|c|}
			\hline
			\multirow{2}{*}{\#FPEM} & \multirow{2}{*}{GFLOPS} & \multicolumn{2}{c|}{ICDAR 2015} & \multicolumn{2}{c|}{CTW1500} \\
			\cline{3-6}
			& & F & FPS & F & FPS \\
			\hline
			0 & \textbf{42.17} & 78.4 & \textbf{33.7} & 78.8 & \textbf{49.7} \\
			\hline
			1 & 42.92 & 79.9 & 29.5 & 80.4 & 44.7 \\
			\hline
			2 & 43.67 & 80.3 & 26.1 & 81.0 & 39.8 \\
			\hline
			3 & 44.43 & 80.4 & 23.0 & 81.3 & 35.2 \\
			\hline
			4 & 45.18 & \textbf{80.5} & 20.1 & \textbf{81.5} & 32.4 \\
			\hline
	\end{tabular}}
	\caption{The results of models with different number of cascaded FPEMs. ``\#FPEM'' means the number of cascaded FPEMs. ``F'' means F-measure.
	The FLOPS are calculated for the input of $640 \times 640 \times 3$.}
	\label{tab:abs_nc}
	\vspace{-4pt}
\end{table}

\textbf{The effectiveness of FPEM.} We design two groups of experiments to verify the effectiveness of FPEM. Firstly, we make a comparison between the model with FPEM and without FPEM.
As shown in Table~\ref{tab:abs_nc}, compared to the model without FPEM ($n_c = 0$), the model with one FPEM ($n_c = 1$) can make about 1.5\% improvement on F-measure while bringing tiny extra computation.
Secondly, we make comparison between a lightweight model equipped with FPEMs and a widely-used segmentation model. To ensure a fair comparison, under the same setting, we employ ``ResNet18 + 2 FPEMs + FFM'' or ``ResNet50 + PSPNet~\cite{zhao2017pyramid}'' as the segmentation network. As shown in Table.~\ref{tab:abs_pspnet}, even the backbone of ``ResNet18 + 2 FPEMs + FFM'' is lightweight, it can reach almost same performance as ``ResNet50 + PSPNet~\cite{zhao2017pyramid}''. In addition, ``ResNet18 + 2 FPEMs + FFM'' enjoy over 5 times faster speed than ``ResNet50 + PSPNet~\cite{zhao2017pyramid}''. The model size of ``ResNet18 + 2 FPEMs + FFM'' is 12.25M.

\begin{table}[t]
	\scriptsize
	\centering
	\renewcommand\arraystretch{1}
	\newcommand{\tabincell}[3]{\begin{tabular}{@{}#1@{}}#2\end{tabular}}
	\scalebox{1}{
		\begin{tabular}{|c|c|c|c|c|}
			\hline
			\multirow{2}{*}{Method} & \multicolumn{2}{c|}{ICDAR 2015} & \multicolumn{2}{c|}{CTW1500} \\
			\cline{2-5}
			& F & FPS & F & FPS \\
			\hline
			ResNet18 + 2 FPEMs + FFM & 80.3 & \textbf{26.1} & 81.0 & \textbf{39.8} \\
			\hline
			ResNet50 + PSPNet~\cite{zhao2017pyramid} & \textbf{80.5} & 4.6 & \textbf{81.1} & 7.1 \\
			\hline
	\end{tabular}}
	\caption{The comparison between ``ResNet18 + 2 FPEMs + FFM'' with ``ResNet50 + PSPNet~\cite{zhao2017pyramid}''. ``F'' means F-measure.}
	\label{tab:abs_pspnet}
\end{table}

\textbf{The effectiveness of FFM.} To investigate the effectiveness of FFM, we firstly remove FFM and concatenate the feature maps in the last feature pyramid $F^{n_c}$ to make final segmentation. The F-measure drop 0.6\%-0.8\% when the FFM is removed (see Table~\ref{tab:abs_many}~\#1 and \#2), which indicates that besides the features from deep layers, the shallow features are also important to semantic segmentation. We then compare FFM with the direct concatenation mentioned in Sec.~\ref{sec:ffm}. The proposed FFM can achieve performance comparable to the direct concatenation (see Table~\ref{tab:abs_many}~\#1 and \#3), while FFM is more efficient.

\textbf{The effectiveness of PA.} We study the validity of PA by removing it from the pipeline. Specifically, we set $\beta$ to 0 in Eqn.~\ref{eqn:loss-tot} in the training phase and merge all neighbor text pixels in step \romannumeral2) of post-processing.
Comparing the method with PA (see Table~\ref{tab:abs_many}~\#1), the F-measure of the model without PA (see Table~\ref{tab:abs_many}~\#4) drops over 1\%, which indicate the effectiveness of PA.

\textbf{The influence of the backbone.} To better analyze the capability of the proposed PAN, we replace the lightweight backbone (ResNet18) to heavier backbone (ResNet50 and VGG16). As shown in Table~\ref{tab:abs_many}~\#5 and \#6, under the same setting, both of ResNet50 and VGG16 can bring over 1\% improvement on ICDAR 2015 and over 0.5\% improvement on CTW1500.
However, the reduction of FPS brought by the heavy backbone is apparent.

\begin{table}[t]
	\scriptsize
	\centering
	\renewcommand\arraystretch{1}
	\newcommand{\tabincell}[3]{\begin{tabular}{@{}#1@{}}#2\end{tabular}}
	\scalebox{1}{
		\begin{tabular}{|c|c|c|c|c|c|c|c|}
			\hline
			\multirow{2}{*}{\#} &
			\multirow{2}{*}{Backbone} &
			\multirow{2}{*}{Fuse} &
			\multirow{2}{*}{PA} &
			\multicolumn{2}{c|}{ICDAR 2015} & \multicolumn{2}{c|}{CTW1500} \\
			\cline{5-8}
			& & & & F & FPS & F & FPS \\
			\hline
			1 & ResNet18 & FFM & \checkmark & 80.3 & 26.1 & 81.0 & 39.8 \\
			\hline
			2 & ResNet18 & - & \checkmark & 79.7 & \textbf{26.2} & 80.2 &  \textbf{40.0} \\
			3& ResNet18 & Concat & \checkmark & 80.4 & 22.3 & 81.2 & 35.9 \\
			\hline
			4 & ResNet18 & FFM & - & 79.3 & 26.1 & 79.8 & 39.9 \\
			\hline
			5 & ResNet50 & FFM & \checkmark & 81.4 & 16.7 & \textbf{81.6} & 26.0 \\
			6 & VGG16 & FFM & \checkmark & \textbf{81.9} & 6.6 & 81.5 & 10.1 \\
			\hline
	\end{tabular}}
	\caption{The results of models with different settings.
	``Fuse'' means the fusion method. ``Concat'' means direct concatenation. ``F'' means F-measure.}
	\label{tab:abs_many}
	\vspace{-4pt}
\end{table}

\subsection{Comparisons with State-of-the-Art Methods}
\textbf{Curve text detection}. To evaluate the performance of our method for detecting curved text instance, we compare the proposed PAN with other state-of-the-art methods on CTW1500 and Total-Text which include many curve text instances.
In the testing phase, we set the short side of images to different scales (320, 512, 640) and evaluate the results using the same evaluation method with \cite{Liu2017Detecting} and \cite{totaltext}.
We report the single-scale performance of PAN on CTW1500 and Total-Text in Table~\ref{tab:ctw1500} and Table~\ref{tab:totaltext}, respectively. Note that the backbone of PAN is set to ResNet18 by default.

On CTW1500, PAN-320 (the short side of input image is 320), without external data pre-training, achieve the F-measure of 77.1\% at an astonishing speed (84.2 FPS), in which the F-measure surpasses most of the counterparts, including the methods with external data pre-training, and the speed is 4 times faster than the fastest method.
When fine-tuning on SynthText pre-trained model, the F-measure of PAN-320 can further be boosted to 79.9\%, and PAN-512 outperform all other methods in F-measure by at least 1.2\% while still keeping nearly real-time speed (58 FPS).

Similar conclusions can be obtained on Total-Text. Without external data pre-training, the speed of PAN-320 is real-time (82.4 FPS)
while the performance is still very competitive (77.1\%), and PAN-640 achieves the F-measure of 83.5\%, surpassing all other state-of-the-art methods (including those with external data) over 0.6\%. With SynthText pre-training, the F-measure of PAN-320 boosting to 79.9\%, and the best F-measure achieve by PAN-640 is 85.0\%, which is 2.1\% better than second-best SPCNet~\cite{spcnet}. Meanwhile, the speed can still maintain nearly 40 FPS.

\begin{figure*}[t]
		\centering
		\setlength{\fboxrule}{0pt}
		\fbox{\includegraphics[width=0.88\textwidth]{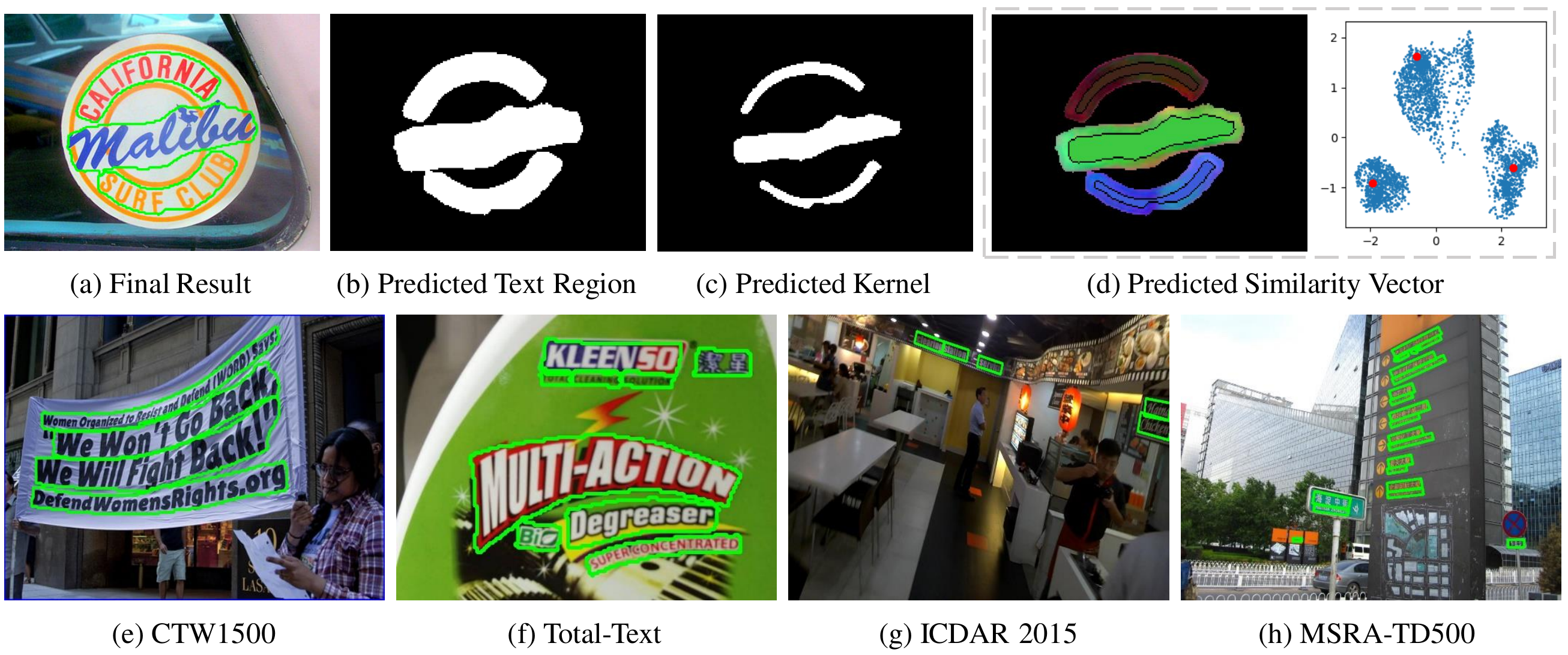}}
		\caption{Qualitative results of PAN. (a) is the final result of PAN. (b) is the predicted text regions. (c) is the predicted kernels. (d) is the visualization of similarity vectors, which is the best viewed in color and scatter diagram. (e)-(h) are results on four standard benchmarks.}
		\label{fig:res}
		\vspace{-4pt}
\end{figure*}

The performance on CTW1500 and Total-Text demonstrates the solid superiority of the proposed PAN to detect arbitrary-shaped text instances. We also illustrate several challenging results in Fig.~\ref{fig:res}~(e)(f), which clearly demonstrate that PAN can elegantly distinguish very complex curve text instances.

\begin{table}[t]
	\scriptsize
	\centering
	\renewcommand\arraystretch{1}
	\newcommand{\tabincell}[3]{\begin{tabular}{@{}#1@{}}#2\end{tabular}}
	\scalebox{0.95}{
		\begin{tabular}{|c|c|c|c|c|c|c|}
			\hline
			\multirow{2}{*}{Method} &
			\multirow{2}{*}{Ext.} &
			\multirow{2}{*}{Venue} & \multicolumn{4}{c|}{CTW1500} \\
			\cline{4-7}
			& & & P & R & F & FPS\\
			\hline
			CTPN*~\cite{tian2016detecting} & - & ECCV'16 & 60.4* & 53.8* & 56.9* & 7.14\\
			\hline
			SegLink*~\cite{shi2017detecting} & - & CVPR'17 & 42.3* & 40.0* & 40.8* & 10.7 \\
			\hline
			EAST*~\cite{zhou2017east} & - & CVPR'17 & 78.7* & 49.1* & 60.4* & 21.2 \\
			\hline
			CTD+TLOC~\cite{Liu2017Detecting} & - & ICDAR'18 & 77.4 & 69.8 & 73.4 & 13.3 \\
			\hline
			PSENet-1s~\cite{psenet} & - & CVPR'19 & 80.6 & 75.6 & 78.0 & 3.9 \\
			\hline
			PAN-320 & - & - & 82.2 & 72.6 & 77.1 & \textbf{84.2}\\
			PAN-512 & - & - & 83.8 & 77.1 & 80.3 & 58.1\\
			PAN-640 & - & - & 84.6 & 77.7 & 81.0 & 39.8\\
			\hline
			\hline
			TextSnake~\cite{textsnake} & \checkmark & ECCV'18 & 67.9 & 85.3 & 75.6 & - \\
			\hline
			PSENet-1s~\cite{psenet} & \checkmark & CVPR'19 & 84.8 & 79.7 & 82.2 & 3.9 \\
			\hline
			PAN-320 & \checkmark & - & 82.7 & 77.4 & 79.9 & \textbf{84.2}\\
			PAN-512 & \checkmark & - & 85.5 & \textbf{81.5} & 83.5 & 58.1\\
			PAN-640 & \checkmark & - & \textbf{86.4} & 81.2 & \textbf{83.7} & 39.8\\
			\hline
	\end{tabular}}
	\caption{The single-scale results on CTW1500. ``P'', ``R'' and ``F'' represent the precision, recall and F-measure respectively. ``Ext.'' indicates external data. * indicates the results from \cite{Liu2017Detecting}.
	}
	\label{tab:ctw1500}
	\vspace{-4pt}
\end{table}

\begin{table}[t]
	\scriptsize
	\centering
	\renewcommand\arraystretch{1}
	\newcommand{\tabincell}[3]{\begin{tabular}{@{}#1@{}}#2\end{tabular}}
	\scalebox{0.95}{
		\begin{tabular}{|c|c|c|c|c|c|c|}
			\hline
			\multirow{2}{*}{Method} & \multirow{2}{*}{Ext.} &
			\multirow{2}{*}{Venue} & \multicolumn{4}{c|}{Total-Text} \\
			\cline{4-7}
			& & & P & R & F & FPS\\
			\hline
			SegLink*~\cite{shi2017detecting} &  & CVPR'17 & 30.3* & 23.8* & 26.7* & - \\
			\hline
			EAST*~\cite{zhou2017east} & - &CVPR'17 & 50.0* & 36.2* & 42.0* & - \\
			\hline
			DeconvNet~\cite{totaltext} & - &ICDAR'18 & 33.0 & 40.0 & 36.0 & - \\
			\hline
			PSENet-1s~\cite{psenet} & - & CVPR'19 & 81.8 & 75.1 & 78.3 & 3.9 \\
			\hline
			PAN-320 & - & - & 84.0 & 71.3 & 77.1 & \textbf{82.4}\\
			PAN-512 & - & - & 86.7 & 78.4 & 82.4 & 57.1\\
			PAN-640 & - & - & 88.0 & 79.4 & 83.5 & 39.6\\
			\hline
			\hline
			TextSnake~\cite{textsnake} & \checkmark & ECCV'18 & 82.7 & 74.5 & 78.4 & - \\
			\hline
			PSENet-1s~\cite{psenet} & \checkmark &CVPR'19 & 84.0 & 78.0 & 80.9 & 3.9 \\
			\hline
			SPCNet~\cite{spcnet} & \checkmark &AAAI'19 & 83.0 & 82.8 & 82.9 & - \\
			\hline
			PAN-320 & \checkmark & - & 85.6 & 75.0 & 79.9 & \textbf{82.4}\\
			PAN-512 & \checkmark & - & \textbf{89.4} & 79.7 & 84.3 & 57.1\\
			PAN-640 & \checkmark & - & 89.3 & \textbf{81.0} & \textbf{85.0} & 39.6\\
			\hline

	\end{tabular}}
	\caption{The single-scale results on Total-Text. ``P'', ``R'' and ``F'' represent the precision, recall and F-measure respectively. ``Ext.'' indicates external data. * indicates the results from \cite{textsnake}.
	}
	\label{tab:totaltext}
\end{table}

\begin{table}[t]
	\scriptsize
	\centering
	\renewcommand\arraystretch{1}
	\newcommand{\tabincell}[3]{\begin{tabular}{@{}#1@{}}#2\end{tabular}}
	\scalebox{0.95}{
		\begin{tabular}{|c|c|c|c|c|c|c|}
			\hline
			\multirow{2}{*}{Method} & \multirow{2}{*}{Ext.} &
			\multirow{2}{*}{Venue}  & \multicolumn{4}{c|}{ICDAR 2015} \\
			\cline{4-7}
			& & & P & R & F & FPS\\
			\hline
			CTPN~\cite{tian2016detecting} &- & ECCV'16 & 74.2 & 51.6 & 60.9 & 7.1 \\
			\hline
			EAST~\cite{zhou2017east} & - & CVPR'17 & 83.6 & 73.5 & 78.2 & 13.2 \\
			\hline
			RRPN~\cite{rrpn} &- & TMM'18 & 82.0 & 73.0 & 77.0 & - \\
			\hline
			DeepReg~\cite{deepreg} & - & ICCV'17 & 82.0 & 80.0 & 81.0 & -\\
			\hline
			PixelLink~\cite{PixelLink} & - &AAAI'18 & 82.9 & 81.7 & 82.3 &7.3\\
			\hline
			PAN & - & - & 82.9 & 77.8 & 80.3 & \textbf{26.1} \\
			\hline
			\hline
			SegLink~\cite{shi2017detecting} &\checkmark &CVPR'17 & 73.1 & 76.8 & 75.0 & - \\
			\hline
			SSTD~\cite{he2017single} &\checkmark &ICCV'17 & 80.2 & 73.9 & 76.9 & 7.7 \\
			\hline
			WordSup~\cite{hu2017wordsup} &\checkmark &CVPR'17 & 79.3 & 77.0 & 78.2 & -  \\
			\hline
			Lyu et al.~\cite{lyu2018multi} & \checkmark &CVPR'18 & \textbf{94.1} & 70.7 & 80.7 & 3.6 \\
			\hline
			RRD~\cite{rrd} &\checkmark &CVPR'18 & 85.6 & 79.0 & 82.2 & 6.5 \\
			\hline
			MCN~\cite{mcn} &\checkmark &CVPR'18 & 72.0 & 80.0 & 76.0 & -\\
			\hline
			TextSnake~\cite{textsnake} & \checkmark &ECCV'18 & 84.9 & 80.4 & 82.6 & 1.1 \\
			\hline
			PSENet-1s~\cite{psenet} & \checkmark &CVPR'19 & 86.9 & 84.5 & 85.7 & 1.6 \\
			\hline
			SPCNet~\cite{spcnet} & \checkmark &AAAI'19 & 88.7 & \textbf{85.8} & \textbf{87.2} & - \\
			\hline
			PAN & \checkmark & - & 84.0 & 81.9 & 82.9 & \textbf{26.1} \\
			\hline

	\end{tabular}}
	\caption{The single-scale results on ICDAR 2015. ``P'', ``R'' and ``F'' represent the precision, recall and F-measure respectively. ``Ext.'' indicates external data.
	}
	\label{tab:ic15}
\end{table}

\begin{table}[t]
	\scriptsize
	\centering
	\renewcommand\arraystretch{1}
	\newcommand{\tabincell}[2]{\begin{tabular}{@{}#1@{}}#2\end{tabular}}
	\scalebox{0.95}{
		\begin{tabular}{|c|c|c|c|c|c|c|c|}
			\hline
			\multirow{2}{*}{Method} & \multirow{2}{*}{Ext.} & \multirow{2}{*}{Venue} & \multicolumn{4}{c|}{MSRA-TD500} \\
			\cline{4-8}
			& & & P & R & F & FPS \\
			\hline
			EAST~\cite{zhou2017east} & - & CVPR'17 & 87.3 & 67.4 & 76.1 & 13.2 \\
			\hline
			RRPN~\cite{rrpn} &- & TMM'18 & 82.0 & 68.0 & 74.0 & - \\
			\hline
			DeepReg~\cite{deepreg} & - & ICCV'17 & 77.0 & 70.0 &74.0 & 1.1\\
			\hline
			PAN & - & - & 80.7 & 77.3 & 78.9 & \textbf{30.2} \\
			\hline
			\hline
			SegLink~\cite{shi2017detecting} & \checkmark & CVPR'17 & 86.0 & 70.0 & 77.0 & 8.9 \\
			\hline
			PixelLink~\cite{PixelLink} & \checkmark & AAAI'18 & 83.0 & 73.2 & 77.8 & 3.0\\
			\hline
			Lyu et al.~\cite{lyu2018multi} & \checkmark & CVPR'18 & 87.6 & 76.2 & 81.5 & 5.7 \\
			\hline
			RRD~\cite{rrd} & \checkmark & CVPR'18 & 87.0 & 73.0 & 79.0 & 10\\
			\hline
			MCN~\cite{mcn} & \checkmark & CVPR'18 & \textbf{88.0} & 79.0 & 83.0 &-\\
			\hline
			TextSnake~\cite{textsnake} & \checkmark & ECCV'18 & 83.2 & 73.9 & 78.3 & 1.1 \\
			\hline
			PAN & \checkmark & - & 84.4 & \textbf{83.8} & \textbf{84.1} & \textbf{30.2} \\
			\hline
	\end{tabular}}
	\caption{The single-scale results on MSRA-TD500. ``P'', ``R'' and ``F'' represent the precision, recall and F-measure respectively. ``Ext.'' indicates external data.
	}
	\label{tab:msra}
	\vspace{-4pt}
\end{table}

\textbf{Oriented text detection}. We evaluate PAN on the ICDAR 2015 to test its ability for oriented text detection. By default, ResNet18 is adopted as the backbone of PAN. During testing, we scale the short side of input images to 736. The comparisons with other state-of-the-art methods are shown in Table~\ref{tab:ic15}. PAN achieves the F-measure of 80.4\% at 26.1 FPS without external data pre-training. Compared with EAST~\cite{zhou2017east}, our method outperforms EAST 2.1\% in F-measure, while the FPS of our method is 2 times of EAST. Fine-tuning on SynthText can further improve the F-measure to 82.9\% which is on par with TextSnake~\cite{textsnake}, but our method can run 25 times faster than TextSnake. Although the performance of our method is not as well as some methods (e.g. PSENet, SPCNet), our method has a least 16 times faster speed (26.1 FPS) than these methods. Some qualitative illustrations are shown in Fig.~\ref{fig:res}~(g). The proposed PAN successfully detects text instances of arbitrary orientations and sizes.


\textbf{Long straight text detection}. To test the robustness of PAN to long straight text instance, we evaluate PAN on MSRA-TD500 benchmark. To ensure fair comparisons, we resize the short edge of test images to 736 as ICDAR 2015. As shown in Table.~\ref{tab:msra}, the proposed PAN achieve F-measures of 78.9\% and 84.1\% when the external data is not used and used respectively. Compared with other state-of-the-art methods, PAN can achieve higher performance and run at a faster speed (30.2 FPS). Thus, PAN is also robust for long straight text detection (see Fig.~\ref{fig:res}~(h)) and can indeed be deployed in complex natural scenarios.

\begin{table}[t]
	\scriptsize
	\centering
	\renewcommand\arraystretch{1}
	\newcommand{\tabincell}[3]{\begin{tabular}{@{}#1@{}}#2\end{tabular}}
	\scalebox{1}{
		\begin{tabular}{|c|c|c|c|c|c|c|}
			\hline
			\multirow{2}{*}{Method} & \multirow{2}{*}{F} & \multicolumn{3}{c|}{Time consumption~(ms)} & \multirow{2}{*}{FPS} \\
			\cline{3-5}
			& & Backbone & Head & Post &\\
			\hline
			PAN-320 & 77.10 & \textbf{4.4} & \textbf{5.4} & \textbf{2.1} & \textbf{84.2} \\
			\hline
			PAN-512 & 80.32 & 6.4 & 7.3 & 3.5 & 58.1 \\
			\hline
			PAN-640 & \textbf{81.00} & 9.8 & 10.1 & 5.2 & 39.8 \\

			\hline
	\end{tabular}}
	\caption{Time consumption of PAN on CTW-1500. The total time consists of backbone, segmentation head and post-processing. ``F'' represents the F-measure.}
	\label{tab:speed}
	\vspace{-4pt}
\end{table}

\subsection{Result Visualization and Speed Analysis}
\textbf{Result visualization.}
An
example of PAN prediction is shown in Fig.~\ref{fig:res}~(a-d). Fig.~\ref{fig:res}~(b) is the predicted text regions which keep the complete shape information of text instances. Fig.~\ref{fig:res}~(c) is the predicted kernels which clearly distinguish different text instances. Fig.~\ref{fig:res}~(d) is a visualization of similarity vectors. The dimensions of these vectors are reduced to 3 and 2 by PCA~\cite{Wold1987Principal} for visualization. We can easily find that pixels belonging to its kernels have similar color and narrow distance with its cluster center~(kernels).

\textbf{Speed analysis.} We specially analyze the time consumption of PAN in different stages. As shown in Table~\ref{tab:speed}, the time costs of backbone and segmentation head are similar, and the time cost of post-processing is half of them. In practical applications, an obvious way to increase speed is to run the network and post-processing in parallel through a basic producer-consumer model, which can reduce the time cost to the original 4/5.
The above experiments are conducted on CTW1500 test set. We evaluate all test images and calculate the average speed. All results in this paper are tested by PyTorch~\cite{pytorch} with batchsize of 1 on one 1080Ti GPU and one 2.20GHz CPU in a single thread.

\section{Conclusion}
In this paper, we have proposed  an efficient framework to detect arbitrary-shaped text in real-time. We firstly introduce a light-weight segmentation head consisting of Feature Pyramid Enhancement Module and Feature Fusion Module, which can benefit the feature extraction while bringing minor extra computation.
Moreover, we propose Pixel Aggregation to predict similarity vectors between text kernels and surrounding pixels.
These two advantages make the PAN become an efficient and accurate arbitrary-shaped text detector.
Extensive experiments on Total-Text and CTW1500 demonstrate the superior advantages
in
speed and
accuracy
when compared to previous state-of-the-art text detectors.

\section*{Acknowledgments}
This work is supported by the Natural Science Foundation of China under Grant 61672273 and Grant 61832008, the Science Foundation for Distinguished Young Scholars of Jiangsu under Grant BK20160021, and Scientific Foundation of State Grid Corporation of China (Research on Ice-wind Disaster Feature Recognition and Prediction by Few-shot Machine Learning in Transmission Lines). Chunhua Shen and his employer received no financial support for the research, authorship, and/or publication of this article.

{\small
\bibliographystyle{ieee_fullname}
\bibliography{egbib}

\begin{thebibliography}{10}\itemsep=-1pt

\bibitem{chen2017rethinking}
Liang-Chieh Chen, George Papandreou, Florian Schroff, and Hartwig Adam.
\newblock Rethinking atrous convolution for semantic image segmentation.
\newblock {\em arXiv preprint arXiv:1706.05587}, 2017.

\bibitem{totaltext}
Chee~Kheng Ch'ng and Chee~Seng Chan.
\newblock Total-text: A comprehensive dataset for scene text detection and
  recognition.
\newblock In {\em Proc. Int. Conf. Document Analysis Recogn.}, 2017.

\bibitem{PixelLink}
Dan Deng, Haifeng Liu, Xuelong Li, and Deng Cai.
\newblock Pixellink: Detecting scene text via instance segmentation.
\newblock In {\em Proc. {AAAI} Conf. Artificial Intell.}, 2018.

\bibitem{deng2009imagenet}
Jia Deng, Wei Dong, Richard Socher, Li-Jia Li, Kai Li, and Li Fei-Fei.
\newblock Imagenet: A large-scale hierarchical image database.
\newblock In {\em Proc. IEEE Conf. Comp. Vis. Patt. Recogn.}, 2009.

\bibitem{fan2018salient}
Deng-Ping Fan, Ming-Ming Cheng, Jiang-Jiang Liu, Shang-Hua Gao, Qibin Hou, and
  Ali Borji.
\newblock Salient objects in clutter: Bringing salient object detection to the
  foreground.
\newblock In {\em Proc. Eur. Conf. Comp. Vis.}, 2018.

\bibitem{fan2017structure}
Deng-Ping Fan, Ming-Ming Cheng, Yun Liu, Tao Li, and Ali Borji.
\newblock Structure-measure: A new way to evaluate foreground maps.
\newblock In {\em Proc. IEEE Int. Conf. Comp. Vis.}, 2017.

\bibitem{fan2018enhanced}
Deng-Ping Fan, Cheng Gong, Yang Cao, Bo Ren, Ming-Ming Cheng, and Ali Borji.
\newblock Enhanced-alignment measure for binary foreground map evaluation.
\newblock {\em arXiv preprint arXiv:1805.10421}, 2018.

\bibitem{fan2019rethinking}
Deng-Ping Fan, Zheng Lin, Jia-Xing Zhao, Yun Liu, Zhao Zhang, Qibin Hou,
  Menglong Zhu, and Ming-Ming Cheng.
\newblock Rethinking rgb-d salient object detection: Models, datasets, and
  large-scale benchmarks.
\newblock {\em arXiv preprint arXiv:1907.06781}, 2019.

\bibitem{fan2019shifting}
Deng-Ping Fan, Wenguan Wang, Ming-Ming Cheng, and Jianbing Shen.
\newblock Shifting more attention to video salient object detection.
\newblock In {\em Proc. IEEE Conf. Comp. Vis. Patt. Recogn.}, 2019.

\bibitem{synthtext}
Ankush Gupta, Andrea Vedaldi, and Andrew Zisserman.
\newblock Synthetic data for text localisation in natural images.
\newblock In {\em Proc. IEEE Conf. Comp. Vis. Patt. Recogn.}, 2016.

\bibitem{he2017mask}
Kaiming He, Georgia Gkioxari, Piotr Doll{\'a}r, and Ross Girshick.
\newblock Mask r-cnn.
\newblock In {\em Proceedings of the IEEE international conference on computer
  vision}, pages 2961--2969, 2017.

\bibitem{maskrcnn}
Kaiming He, Georgia Gkioxari, Piotr Doll{\'a}r, and Ross Girshick.
\newblock {Mask R-CNN}.
\newblock In {\em Proc. IEEE Int. Conf. Comp. Vis.}, pages 2961--2969, 2017.

\bibitem{he2015delving}
Kaiming He, Xiangyu Zhang, Shaoqing Ren, and Jian Sun.
\newblock Delving deep into rectifiers: Surpassing human-level performance on
  imagenet classification.
\newblock In {\em Proc. IEEE Int. Conf. Comp. Vis.}, 2015.

\bibitem{he2016deep}
Kaiming He, Xiangyu Zhang, Shaoqing Ren, and Jian Sun.
\newblock Deep residual learning for image recognition.
\newblock In {\em Proc. IEEE Conf. Comp. Vis. Patt. Recogn.}, 2016.

\bibitem{he2016identity}
Kaiming He, Xiangyu Zhang, Shaoqing Ren, and Jian Sun.
\newblock Identity mappings in deep residual networks.
\newblock In {\em Proc. Eur. Conf. Comp. Vis.}, 2016.

\bibitem{he2017single}
Pan He, Weilin Huang, Tong He, Qile Zhu, Yu Qiao, and Xiaolin Li.
\newblock Single shot text detector with regional attention.
\newblock In {\em Proc. IEEE Int. Conf. Comp. Vis.}, 2017.

\bibitem{deepreg}
Wenhao He, Xu-Yao Zhang, Fei Yin, and Cheng-Lin Liu.
\newblock Deep direct regression for multi-oriented scene text detection.
\newblock In {\em Proc. IEEE Int. Conf. Comp. Vis.}, 2017.

\bibitem{howard2017mobilenets}
Andrew~G Howard, Menglong Zhu, Bo Chen, Dmitry Kalenichenko, Weijun Wang,
  Tobias Weyand, Marco Andreetto, and Hartwig Adam.
\newblock Mobilenets: Efficient convolutional neural networks for mobile vision
  applications.
\newblock {\em arXiv preprint arXiv:1704.04861}, 2017.

\bibitem{hu2017wordsup}
Han Hu, Chengquan Zhang, Yuxuan Luo, Yuzhuo Wang, Junyu Han, and Errui Ding.
\newblock Wordsup: Exploiting word annotations for character based text
  detection.
\newblock In {\em Proc. IEEE Int. Conf. Comp. Vis.}, 2017.

\bibitem{huang2017densely}
Gao Huang, Zhuang Liu, Kilian~Q Weinberger, and Laurens van~der Maaten.
\newblock Densely connected convolutional networks.
\newblock In {\em Proc. IEEE Conf. Comp. Vis. Patt. Recogn.}, 2017.

\bibitem{ioffe2015batch}
Sergey Ioffe and Christian Szegedy.
\newblock Batch normalization: Accelerating deep network training by reducing
  internal covariate shift.
\newblock {\em arXiv preprint arXiv:1502.03167}, 2015.

\bibitem{karatzas2015icdar}
Dimosthenis Karatzas, Lluis Gomez-Bigorda, Anguelos Nicolaou, Suman Ghosh,
  Andrew Bagdanov, Masakazu Iwamura, Jiri Matas, Lukas Neumann,
  Vijay~Ramaseshan Chandrasekhar, Shijian Lu, et~al.
\newblock Icdar 2015 competition on robust reading.
\newblock In {\em Proc. Int. Conf. Document Analysis Recogn.}, 2015.

\bibitem{lecun1998gradient}
Yann LeCun, L{\'e}on Bottou, Yoshua Bengio, and Patrick Haffner.
\newblock Gradient-based learning applied to document recognition.
\newblock {\em Proceedings of the IEEE}, 1998.

\bibitem{psenet}
Xiang Li, Wenhai Wang, Wenbo Hou, Ruo-Ze Liu, Tong Lu, and Jian Yang.
\newblock Shape robust text detection with progressive scale expansion network.
\newblock {\em arXiv preprint arXiv:1806.02559}, 2018.

\bibitem{li2019selective}
Xiang Li, Wenhai Wang, Xiaolin Hu, and Jian Yang.
\newblock Selective kernel networks.
\newblock In {\em Proc. IEEE Conf. Comp. Vis. Patt. Recogn.}, 2019.

\bibitem{textboxes++}
Minghui Liao, Baoguang Shi, and Xiang Bai.
\newblock Textboxes++: A single-shot oriented scene text detector.
\newblock {\em {IEEE} Trans. Image Process.}, 2018.

\bibitem{liao2017textboxes}
Minghui Liao, Baoguang Shi, Xiang Bai, Xinggang Wang, and Wenyu Liu.
\newblock Textboxes: A fast text detector with a single deep neural network.
\newblock In {\em Proc. {AAAI} Conf. Artificial Intell.}, 2017.

\bibitem{rrd}
Minghui Liao, Zhen Zhu, Baoguang Shi, Gui-song Xia, and Xiang Bai.
\newblock Rotation-sensitive regression for oriented scene text detection.
\newblock In {\em Proc. IEEE Conf. Comp. Vis. Patt. Recogn.}, 2018.

\bibitem{lin2017feature}
Tsung-Yi Lin, Piotr Doll{\'a}r, Ross Girshick, Kaiming He, Bharath Hariharan,
  and Serge Belongie.
\newblock Feature pyramid networks for object detection.
\newblock In {\em Proc. IEEE Conf. Comp. Vis. Patt. Recogn.}, 2017.

\bibitem{liu2016ssd}
Wei Liu, Dragomir Anguelov, Dumitru Erhan, Christian Szegedy, Scott Reed,
  Cheng-Yang Fu, and Alexander~C Berg.
\newblock Ssd: Single shot multibox detector.
\newblock In {\em Proc. Eur. Conf. Comp. Vis.}, 2016.

\bibitem{Liu2017Detecting}
Yuliang Liu, Lianwen Jin, Shuaitao Zhang, and Sheng Zhang.
\newblock Detecting curve text in the wild: New dataset and new solution.
\newblock 2017.

\bibitem{mcn}
Zichuan Liu, Guosheng Lin, Sheng Yang, Jiashi Feng, Weisi Lin, and Wang~Ling
  Goh.
\newblock Learning markov clustering networks for scene text detection.
\newblock {\em Proc. IEEE Conf. Comp. Vis. Patt. Recogn.}, 2018.

\bibitem{long2015fully}
Jonathan Long, Evan Shelhamer, and Trevor Darrell.
\newblock Fully convolutional networks for semantic segmentation.
\newblock In {\em Proceedings of the IEEE conference on computer vision and
  pattern recognition}, pages 3431--3440, 2015.

\bibitem{FCN}
Jonathan Long, Evan Shelhamer, and Trevor Darrell.
\newblock Fully convolutional networks for semantic segmentation.
\newblock In {\em Proc. IEEE Conf. Comp. Vis. Patt. Recogn.}, 2015.

\bibitem{textsnake}
Shangbang Long, Jiaqiang Ruan, Wenjie Zhang, Xin He, Wenhao Wu, and Cong Yao.
\newblock Textsnake: A flexible representation for detecting text of arbitrary
  shapes.
\newblock {\em Proc. Eur. Conf. Comp. Vis.}, 2018.

\bibitem{masttextspotter}
Pengyuan Lyu, Minghui Liao, Cong Yao, Wenhao Wu, and Xiang Bai.
\newblock Mask textspotter: An end-to-end trainable neural network for spotting
  text with arbitrary shapes.
\newblock In {\em Proc. Eur. Conf. Comp. Vis.}, 2018.

\bibitem{lyu2018multi}
Pengyuan Lyu, Cong Yao, Wenhao Wu, Shuicheng Yan, and Xiang Bai.
\newblock Multi-oriented scene text detection via corner localization and
  region segmentation.
\newblock {\em arXiv preprint arXiv:1802.08948}, 2018.

\bibitem{rrpn}
Jianqi Ma, Weiyuan Shao, Hao Ye, Li Wang, Hong Wang, Yingbin Zheng, and
  Xiangyang Xue.
\newblock Arbitrary-oriented scene text detection via rotation proposals.
\newblock {\em IEEE Transactions on Multimedia}, 2018.

\bibitem{milletari2016v}
Fausto Milletari, Nassir Navab, and Seyed-Ahmad Ahmadi.
\newblock V-net: Fully convolutional neural networks for volumetric medical
  image segmentation.
\newblock In {\em Proc. Int. Conf. 3D Vision}, 2016.

\bibitem{pytorch}
Adam Paszke, Sam Gross, Soumith Chintala, Gregory Chanan, Edward Yang, Zachary
  DeVito, Zeming Lin, Alban Desmaison, Luca Antiga, and Adam Lerer.
\newblock Automatic differentiation in pytorch.
\newblock 2017.

\bibitem{ren2015faster}
Shaoqing Ren, Kaiming He, Ross Girshick, and Jian Sun.
\newblock Faster r-cnn: Towards real-time object detection with region proposal
  networks.
\newblock In {\em Proc. Advances in Neural Inf. Process. Syst.}, 2015.

\bibitem{shi2017detecting}
Baoguang Shi, Xiang Bai, and Serge Belongie.
\newblock Detecting oriented text in natural images by linking segments.
\newblock In {\em Proc. IEEE Conf. Comp. Vis. Patt. Recogn.}, 2017.

\bibitem{shrivastava2016training}
Abhinav Shrivastava, Abhinav Gupta, and Ross Girshick.
\newblock Training region-based object detectors with online hard example
  mining.
\newblock In {\em Proc. IEEE Conf. Comp. Vis. Patt. Recogn.}, 2016.

\bibitem{simonyan2014very}
Karen Simonyan and Andrew Zisserman.
\newblock Very deep convolutional networks for large-scale image recognition.
\newblock In {\em Proc. Int. Conf. Learn. Representations}, 2015.

\bibitem{sutskever2013importance}
Ilya Sutskever, James Martens, George Dahl, and Geoffrey Hinton.
\newblock On the importance of initialization and momentum in deep learning.
\newblock In {\em ICML}, 2013.

\bibitem{tang2018cu}
Zhiqiang Tang, Xi Peng, Shijie Geng, Yizhe Zhu, and Dimitris Metaxas.
\newblock Cu-net: Coupled u-nets.
\newblock In {\em BMVC}, 2018.

\bibitem{tian2016detecting}
Zhi Tian, Weilin Huang, Tong He, Pan He, and Yu Qiao.
\newblock Detecting text in natural image with connectionist text proposal
  network.
\newblock In {\em Proc. Eur. Conf. Comp. Vis.}, 2016.

\bibitem{wang2018mixed}
Wenhai Wang, Xiang Li, Tong Lu, and Jian Yang.
\newblock Mixed link networks.
\newblock In {\em Proc. Int. Joint Conf. Artificial Intell.}, 2018.

\bibitem{Wold1987Principal}
Svante Wold, Kim Esbensen, and Paul Geladi.
\newblock Principal component analysis.
\newblock {\em Chemometrics and Intelligent Laboratory Systems}, 2(1-3):37--52,
  1987.

\bibitem{spcnet}
Enze Xie, Yuhang Zang, Shuai Shao, Gang Yu, Cong Yao, and Guangyao Li.
\newblock Scene text detection with supervised pyramid context network.
\newblock In {\em Proc. {AAAI} Conf. Artificial Intell.}, 2019.

\bibitem{yao2014unified}
Cong Yao, Xiang Bai, and Wenyu Liu.
\newblock A unified framework for multioriented text detection and recognition.
\newblock {\em IEEE Transactions on Image Processing}, 23(11):4737--4749, 2014.

\bibitem{msra}
Cong Yao, Xiang Bai, Wenyu Liu, Yi Ma, and Zhuowen Tu.
\newblock Detecting texts of arbitrary orientations in natural images.
\newblock In {\em Proc. IEEE Conf. Comp. Vis. Patt. Recogn.}, 2012.

\bibitem{yao2016scene}
Cong Yao, Xiang Bai, Nong Sang, Xinyu Zhou, Shuchang Zhou, and Zhimin Cao.
\newblock Scene text detection via holistic, multi-channel prediction.
\newblock {\em arXiv preprint arXiv:1606.09002}, 2016.

\bibitem{yu2018bisenet}
Changqian Yu, Jingbo Wang, Chao Peng, Changxin Gao, Gang Yu, and Nong Sang.
\newblock Bisenet: Bilateral segmentation network for real-time semantic
  segmentation.
\newblock In {\em Proc. Eur. Conf. Comp. Vis.}, 2018.

\bibitem{zhang2016multi}
Zheng Zhang, Chengquan Zhang, Wei Shen, Cong Yao, Wenyu Liu, and Xiang Bai.
\newblock Multi-oriented text detection with fully convolutional networks.
\newblock In {\em Proc. IEEE Conf. Comp. Vis. Patt. Recogn.}, 2016.

\bibitem{zhao2017pyramid}
Hengshuang Zhao, Jianping Shi, Xiaojuan Qi, Xiaogang Wang, and Jiaya Jia.
\newblock Pyramid scene parsing network.
\newblock In {\em Proc. IEEE Conf. Comp. Vis. Patt. Recogn.}, 2017.

\bibitem{zhao2019contrast}
Jia-Xing Zhao, Yang Cao, Deng-Ping Fan, Ming-Ming Cheng, Xuan-Yi Li, and Le
  Zhang.
\newblock Contrast prior and fluid pyramid integration for rgbd salient object
  detection.
\newblock In {\em Proc. IEEE Conf. Comp. Vis. Patt. Recogn.}, 2019.

\bibitem{zhou2017east}
Xinyu Zhou, Cong Yao, He Wen, Yuzhi Wang, Shuchang Zhou, Weiran He, and Jiajun
  Liang.
\newblock East: an efficient and accurate scene text detector.
\newblock {\em arXiv preprint arXiv:1704.03155}, 2017.

\end{thebibliography}
}
\clearpage
\section{Appendix}

\maketitle
\ificcvfinal\thispagestyle{empty}\fi

\subsection{Robustness Analysis}
To further demonstrate the robustness of the proposed PAN, we evaluate the model by training on one dataset and testing on other datasets.
Based on the annotation level, we divide the datasets into two groups which are word level and text line level datasets.
SynthText, ICDAR 2015 and Total-Text are annotated at word level, while CTW1500 and MSRA-TD500 are annotated at text line level.
For fair comparisons, we train all model without any external dataset, and the short side of test images in ICDAR 2015, MSRA-TD500, CTW1500 and Total-Text are set to 736, 736, 640, 640 respectively.

The cross-dataset results of PAN are shown in Table~\ref{tab:cross}.
Notably, the proposed PAN trained on SynthText (a synthetic dataset) have reluctantly satisfied performance on ICDAR 2015 and Total-Text, which indicates that even without any manually annotated data, PAN can satisfy the scene with low precision requirements.
The PAN trained on manually annotated dataset has over 64\% F-measure in the cross-dataset evaluation, which is still competitive.
Furthermore, in the cross-dataset evaluation at text line level, all models achieve the F-measure of nearly 75\% even the training and the testing are performed on quadrangle and curved text datasets respectively. These cross-dataset experiments demonstrate that the proposed PAN is robust in generalizing to brand new datasets.


\begin{table}[b]
	\scriptsize
	\centering
	\renewcommand\arraystretch{1}
	\newcommand{\tabincell}[3]{\begin{tabular}{@{}#1@{}}#2\end{tabular}}
	\scalebox{1.1}{
		\begin{tabular}{|c|c|c|c|c|}
			\hline
			Annotation & Train Set$\rightarrow$Test Set & P & R & F\\
			\hline
			\multirow{4}{*}{Word} & SynthText~$\rightarrow$~ICDAR 2015 & 65.9 & 46.9 & 54.8\\
			& SynthText~$\rightarrow$~Total-Text & 69.1 & 40.8 & 51.3 \\
			& ICDAR 2015~$\rightarrow$~Total-Text & 72.0 & 57.8 & 64.1\\
			& Total-Text~$\rightarrow$~ICDAR 2015 & 77.6 & 65.5 & 71.1\\
			\hline
			\multirow{2}{*}{Text Line} & CTW1500~$\rightarrow$~MSRA-TD500 & 76.6 & 73.1 & 74.8\\
			& MSRA-TD500~$\rightarrow$~CTW1500 & 82.4 & 69.1 & 75.2\\
			\hline
	\end{tabular}}
	\caption{Cross-dataset results of PAN on word-level and line-level datasets. ``P'', ``R'' and ``F'' represent the precision, recall and F-measure respectively.
	}
	\label{tab:cross}
\end{table}

\subsection{Comparisons with Other Semantic Segmentation Methods}
Unlike common semantic segmentation tasks, text detection needs to distinguish different text instances that lie closely.
So feature map resolution matters and cannot be too small.
However, most of high efficiency segmentation methods (i.e. BiSeNet~\cite{yu2018bisenet}) make prediction on $1/8$ feature map, sacrificing accuracy for speed. Their speed will reduce sharply if using $1/4$ feature map directly.
Thus, `how to keep the high efficiency and the high resolution feature map simultaneously?'' is a challenging problem, and our answer is ``ResNet18 + 2FPEM + FFM''.
We compare our method with two methods BiSeNet~\cite{yu2018bisenet} and CU-Net~\cite{tang2018cu} on CTW1500.
For fair comparisons, we set the backbone of BiSeNet to ResNet18 and use one of default settings of CU-Net, which has the similar speed with our method.
As shown in Table~\ref{tab:diff_seg}, our method enjoys obviously better accuracy (+2.2\% and +4.6\%) at the similar speed.

\begin{table}[t]
	\scriptsize
	\centering
	\renewcommand\arraystretch{1.0}
	\newcommand{\tabincell}[2]{\begin{tabular}{@{}#1@{}}#2\end{tabular}}
	\scalebox{1.1}{
		\begin{tabular}{|c|c|c|c|}
			\hline
			Methods & Ext. & F (\%) & FPS \\
			\hline
			BiSeNet (ResNet18)~\cite{yu2018bisenet} & - & 78.8 & 25.9 \\ 
			\hline
			CU-Net-2 ($m$=128, $n$=32)~\cite{tang2018cu} & - & 76.4 & 39.3 \\
			\hline
			Ours (ResNet18 + 2FPEM + FFM) & - & \textbf{81.0} & \textbf{39.8} \\
			\hline
	\end{tabular}}
	\caption{The results on CTW1500 of different segmentation methods. 
	``F'' means F-measure. ``Ext.'' indicates external data.}
	\label{tab:diff_seg}
	\vspace{-10pt}
\end{table}

\begin{figure}[t]
	\centering
	\setlength{\fboxrule}{0pt}
	\fbox{\includegraphics[width=0.4\textwidth]{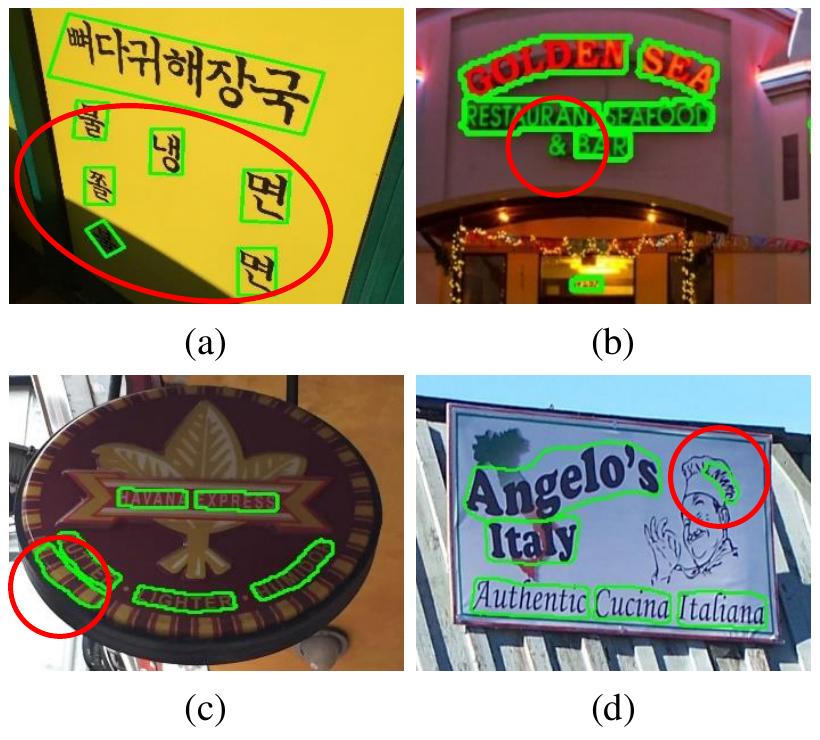}}
	\caption{Failure Samples.}
	\label{fig:failure}
\end{figure}

\subsection{Failure Samples}
As demonstrated in previous experiments, the proposed PAN works well in most cases of arbitrary-shaped text detection.
It still fails for some difficult cases, such as large character spacing (see Fig.~\ref{fig:failure}~(a)), symbols (see Fig.~\ref{fig:failure}~(b)) and false positives (see Fig.~\ref{fig:failure}~(c)(d)).
Large character spacing is an unresolved problem which also exists in other state-of-the-art methods such as RRD~\cite{rrd}.
For symbol detection and false positives, PAN is trained on small datasets (about 1000 images) and we believe this problem will be alleviated when increasing training data.


\begin{figure*}[b]
	\centering
	\setlength{\fboxrule}{0pt}
	\fbox{\includegraphics[width=0.9\textwidth]{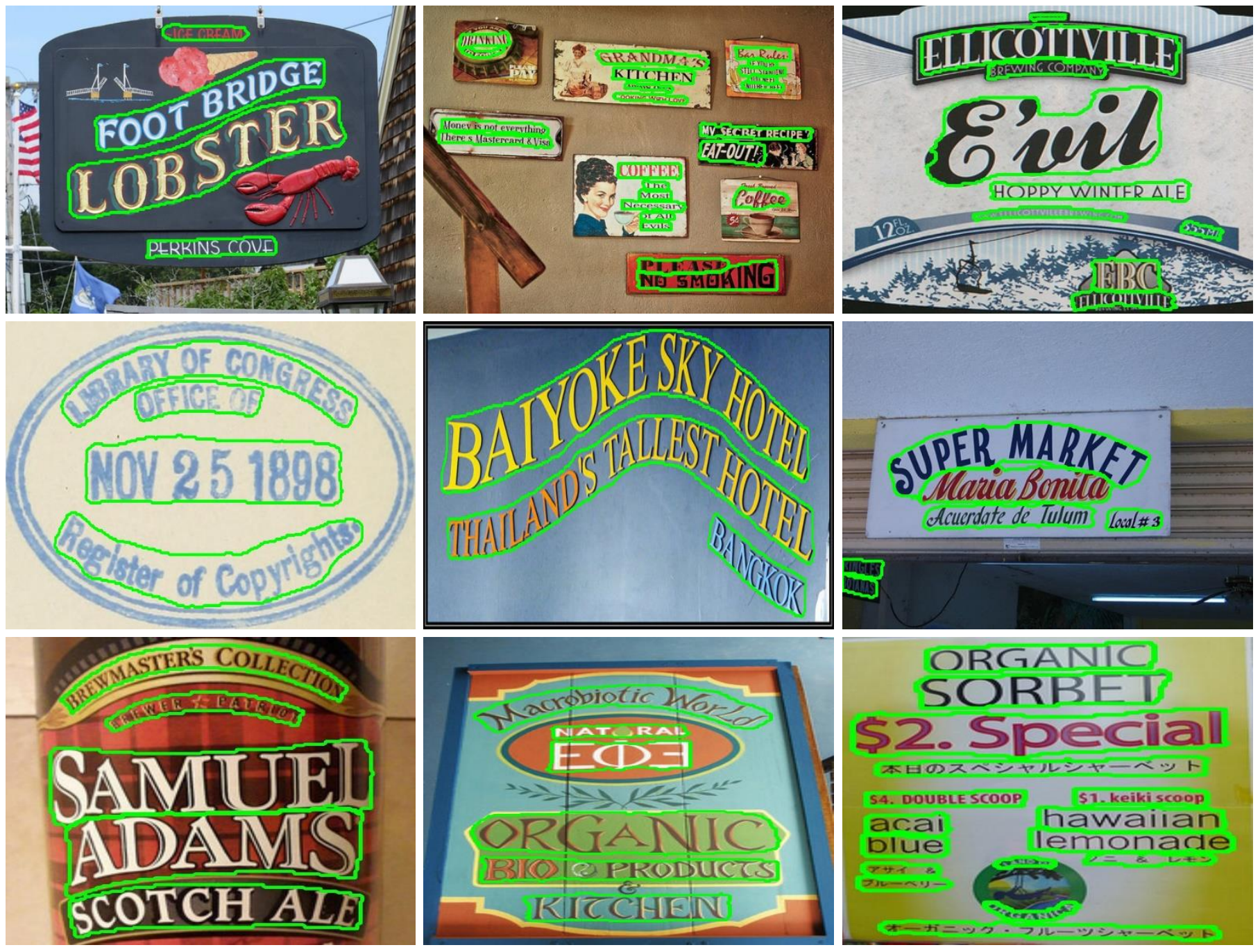}}
	\caption{Detection results on CTW1500.}
	\label{fig:ctw}
\end{figure*}

\begin{figure*}[b]
	\centering
	\setlength{\fboxrule}{0pt}
	\fbox{\includegraphics[width=0.9\textwidth]{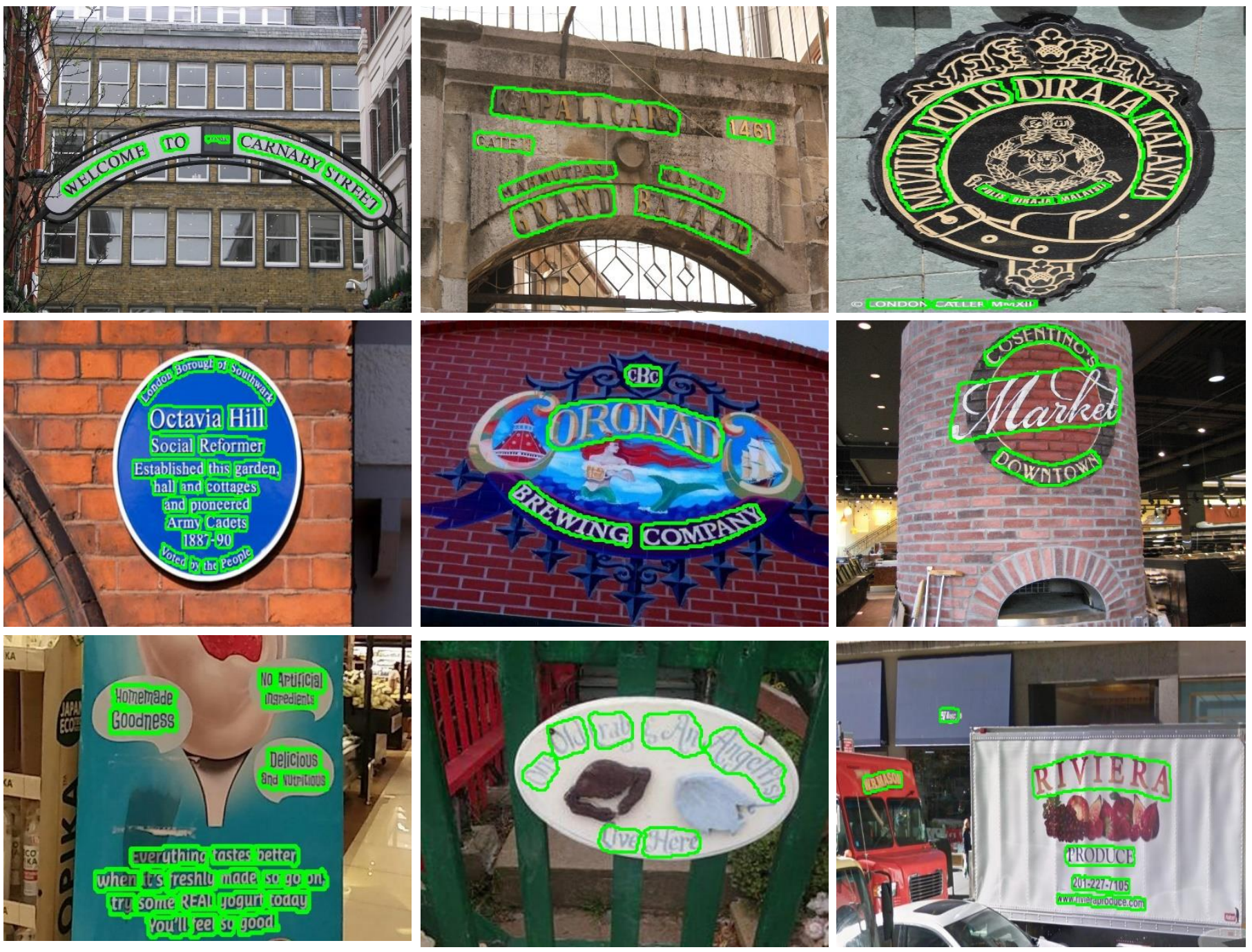}}
	\caption{Detection results on Total-Text.}
	\label{fig:tt}
\end{figure*}

\begin{figure*}[b]
	\centering
	\setlength{\fboxrule}{0pt}
	\fbox{\includegraphics[width=0.9\textwidth]{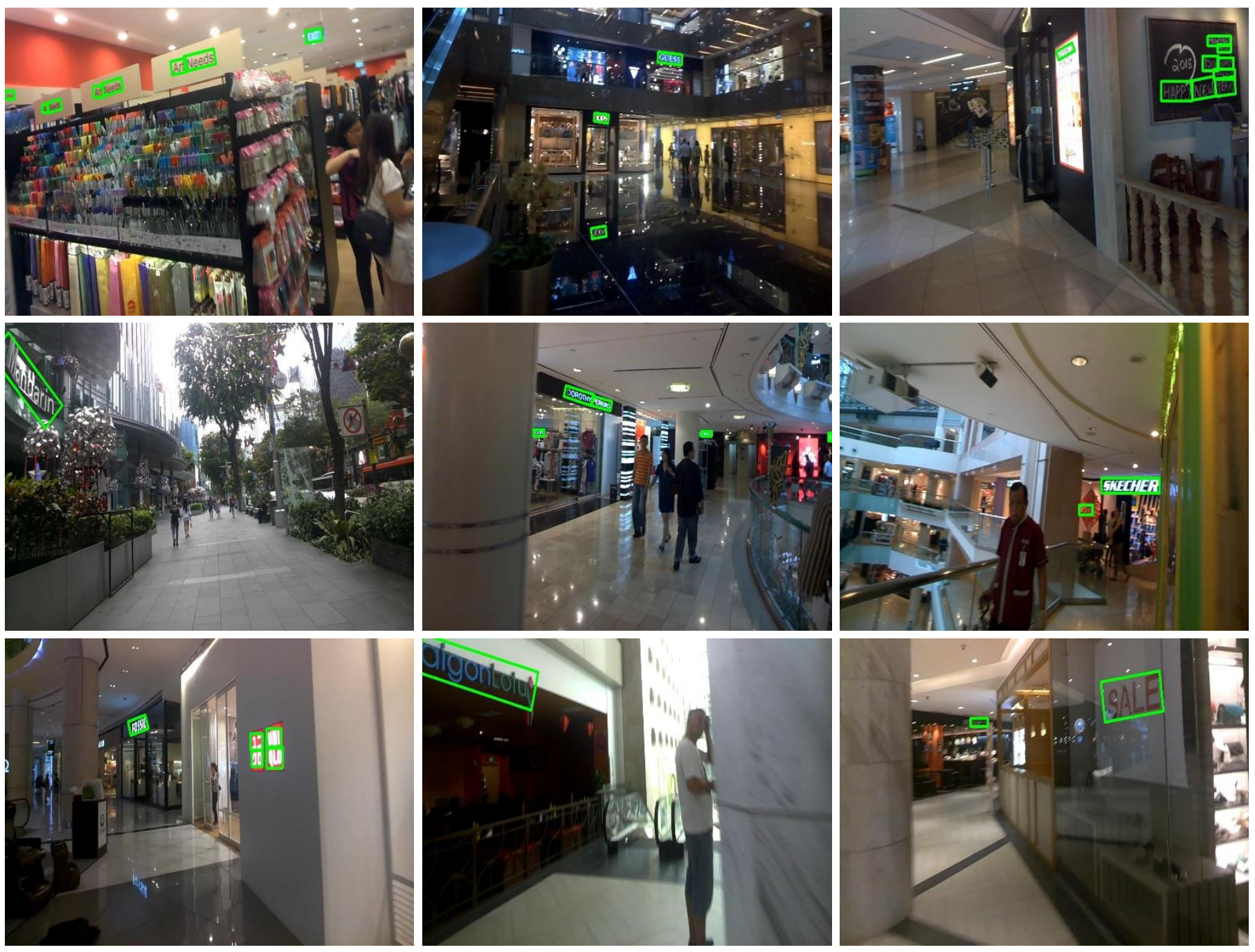}}
	\caption{Detection results on ICDAR 2015.}
	\label{fig:ic15}
\end{figure*}

\begin{figure*}[b]
	\centering
	\setlength{\fboxrule}{0pt}
	\fbox{\includegraphics[width=0.9\textwidth]{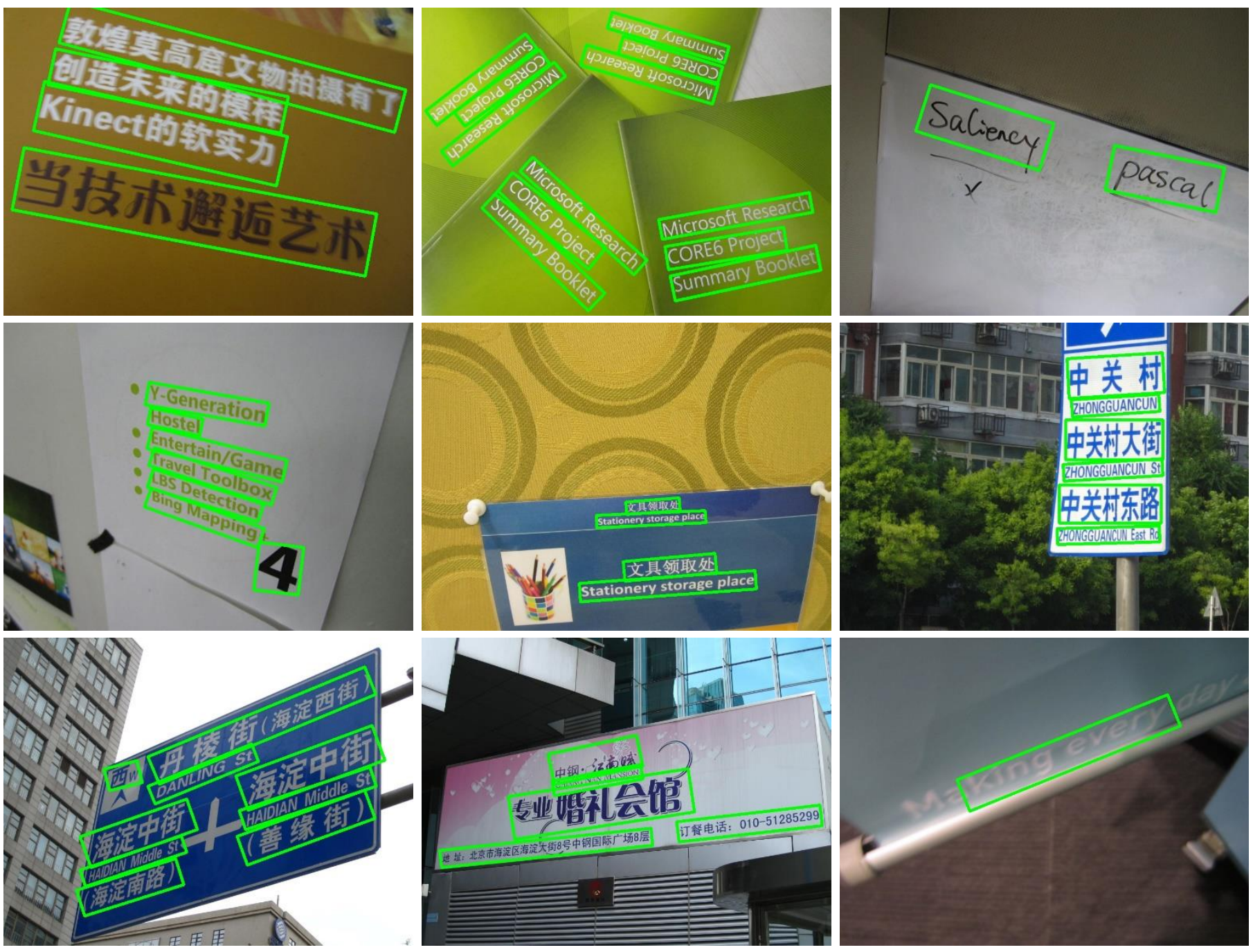}}
	\caption{Detection results on MSRA-TD500.}
	\label{fig:msra}
\end{figure*}

\subsection{More Detected Results on CTW1500, Total Text, ICDAR 2015 and MSRA-TD500}
In this section, we show more test examples produced by PAN on different datasets in Fig.~\ref{fig:ctw} (CTW1500) Fig.~\ref{fig:tt} (Total-Text), Fig.~\ref{fig:ic15} (ICDAR 2015) and Fig.~\ref{fig:msra} (MSRA-TD500).
From these results, we can find that the proposed PAN have the following abilities:
\romannumeral1) separating adjacent text instances with narrow distances;
\romannumeral2) locating the arbitrary-shaped text instances precisely;
\romannumeral3) detecting the text instances with various orientations;
\romannumeral4) detecting the long text instances;
\romannumeral5) detecting the multiple Lingual text.
Meanwhile, thanks to the strong feature representation, PAN can also locate the text instances with complex and unstable illumination, different colors and variable scales.
\end{document}